# Machine Learning for Sentiment Analysis of Imported Food in Trinidad and Tobago

**Authors:** Cassandra Daniels and Koffka Khan

**Abstract**

This research investigates the performance of various machine learning algorithms (CNN, LSTM, VADER, and RoBERTa) for sentiment analysis of Twitter data related to imported food items in Trinidad and Tobago. The study addresses three primary research questions: the comparative accuracy and efficiency of the algorithms, the optimal configurations for each model, and the potential applications of the optimized models in a live system for monitoring public sentiment and its impact on the import bill. The dataset comprises tweets from 2018 to 2024, divided into imbalanced, balanced, and temporal subsets to assess the impact of data balancing and the COVID-19 pandemic on sentiment trends. Ten experiments were conducted to evaluate the models under various configurations. Results indicated that VADER outperformed the other models in both multi-class and binary sentiment classifications. The study highlights significant changes in sentiment trends pre- and post-COVID-19, with implications for import policies.

## Section 1: Introduction

### 1.1 Background of the Problem

The rise of social media has not only changed personal communication but also transformed the landscape of business and governance. Platforms like Twitter offer a real-time, unfiltered glimpse into public opinion, making it a valuable resource for sentiment analysis. In Trinidad and Tobago, the reliance on imported food items is a critical aspect of national policy due to the country's geographical and agricultural limitations. The ability to gauge public sentiment towards these imports can help address issues related to consumer satisfaction, economic dependency, and food security. Traditionally, sentiment analysis has been used in various domains such as finance, marketing, and politics. However, its application in understanding sentiments related to food imports is relatively novel. This study aims to bridge this gap by focusing on Twitter data, which is abundant and provides a broad spectrum of opinions. By leveraging machine learning (ML) algorithms, this research seeks to provide a deeper understanding of public sentiment and its potential influence on import policies.

### 1.2 Statement of the Problem

In the context of Trinidad and Tobago, the food import sector is not only an economic concern but also a matter of public interest and national security. Despite this, there is a noticeable gap in research that specifically targets sentiment analysis of public opinions on imported food. Previous studies have primarily concentrated on general sentiment analysis, leaving a significant research void in this specific and impactful area. Moreover, the dynamic and unstructured nature of Twitter data presents additional challenges for sentiment analysis. Traditional methods often fall short in handling the nuances of informal language, slang, and the brevity of tweets.

This necessitates the use of advanced ML algorithms that can effectively manage these challenges and provide accurate sentiment classifications. The onset of the COVID-19 pandemic has further complicated this landscape, influencing public sentiment in unprecedented ways. The pandemic has disrupted global supply chains, altered consumer behaviours , and heightened concerns about food security. Understanding how these factors have impacted public sentiment towards imported food is essential for developing responsive and effective policies.

### 1.3 Purpose of the Study

The study's comprehensive approach aims not only to evaluate the performance of different ML algorithms but also to understand the underlying sentiment trends. By analysing a dataset of tweets over an extended period (2018-2024), this research provides a longitudinal view of public opinion, capturing shifts and trends that may be associated with significant events such as the COVID-19 pandemic. The optimization of model configurations is another crucial aspect of this study. ML models often require fine-tuning to achieve optimal performance. This research systematically explores various configurations for each algorithm, identifying the best practices for sentiment analysis in this context. The goal is to develop robust models that can be deployed in real-time systems for continuous monitoring of public sentiment. Furthermore, the study explores practical applications of these models in a live monitoring system. By doing so, it seeks to demonstrate the feasibility and utility of using sentiment analysis for decision-making in the food import sector. The insights gained from this analysis can help policymakers and businesses make informed decisions that are aligned with public sentiment, potentially improving public satisfaction and economic outcomes.

### 1.4 Research Questions

This research is guided by the following questions:

1. How do the CNN, LSTM, VADER, and RoBERTa algorithms compare in terms of accuracy and efficiency for sentiment analysis of Twitter data related to imported food in Trinidad and Tobago?
2. What are the optimal configurations for each of these models to achieve the best performance?
3. How can the optimized models be applied in a live system to monitor public sentiment, and what implications do the sentiment trends have on the import bill of Trinidad and Tobago?

### 1.5 Significance of the Study

The results of this study have significant implications. For policymakers, the ability to track and understand public sentiment towards imported food can lead to more responsive and effective policies. This is particularly crucial in Trinidad and Tobago, where food imports are a major part of the national economy. By aligning policies with public sentiment, policymakers can enhance public trust and satisfaction, potentially leading to more sustainable economic outcomes. For businesses in the food import sector, sentiment analysis offers valuable market intelligence. Understanding consumer preferences and concerns can inform product offerings, marketing strategies, and customer engagement efforts. This can result in better business decisions, improved customer satisfaction, and increased market competitiveness. For researchers, this study contributes to the growing body of knowledge in sentiment analysis and machine learning. It provides a comparative evaluation of different algorithms in a specific and practical context, highlighting their strengths, weaknesses, and best practices. The insights gained from this research can guide future studies and applications in other fields. The study also explores the impact of the COVID-19 pandemic on public sentiment, offering a unique perspective on how global events can influence local opinions and behaviours. By analysing sentiment trends over time, this research provides valuable insights into the dynamic nature of public opinion and its implications for policy and business decisions.

## Section 2: Literature Review

Sentiment Analysis (SA) analyses text in order to identify peoples' opinions, attitudes and emotions in relation to any entity to determine the sentiment they express and classify their polarity (Medhat 2014). This classification process has levels such as document-level, sentence-level, phrase-level and aspect-level SA (Wankhade 2022). The document-level considers a document as a single unit of information and then classifies it as expressing positive or negative sentiment (Parminder Bhatia 2015). A determination on whether a sentence is subjective or objective is carried out by the sentence -level SA followed by if it portrays positive or negative sentiments if subjective (Bishan Yang 2014). Phrase-level SA focuses on mining opinion words within phrases and then classifying them. Each phrase can contain one or multiple aspects, which is particularly useful for product reviews spanning multiple lines where individual phrases express distinct aspects (Zhang 2015). While document-level analysis categorizes an entire document as positive or negative, and sentence-level analysis identifies sentiment within sentences, phrase-level analysis digs deeper into the finer details of sentiment expressed in smaller textual units. Aspect-level SA focuses on identifying the sentiment towards specific aspects or features within the text. While overall sentiment analysis refers to the sentiment towards an entire entity, aspect-level sentiment analysis focuses on the sentiments associated with specific aspects of the entity being discussed. This approach allows for a more detailed and nuanced analysis, utilizing more of the information provided by the textual review (Kim Schouten 2015). It is crucial to acknowledge that sentiment analysis has its limitations. Social media content frequently includes sarcasm, irony, slang, and nuanced language, which can be difficult for automated algorithms to accurate interpret. Additionally, cultural differences, context, and linguistic variations across languages add further complexity to the analysis (Sande 2024).

The advancements in technology has allowed persons to see the world through multiple lenses as it has significantly broadened the range of information and perspectives available to people. Through these lenses persons can make more informed opinions on various topics such as global issues, news and current events, staying connected socially, entertainment and culture, consumer choices and more. Product reviews and ratings are integral tools that help consumers make informed purchasing decisions. Online retailers also benefit from these tools as they use rating systems to build trust and establish a good reputation in the e-commerce market. Many online stores provide quantitative ratings, textual reviews, or a mix of both (Georg Lackermair 2013). The utilization of these reviews can be seen in many different industries, for example in the food industry, companies may adjust flavours or ingredients in specific locations to align with regional preferences indicated by reviews, leveraging popular terms or phrases from rating scores (Md Shodel Sayeed 2023). The literature also reflects that persons' opinions can cause shifts in the stock market. Positive news have the power to drive investors to purchase shares, causing share prices to rise, while negative news can have the opposite effect. Sentiment classification of the stock market has been a promising method for predicting stock trends by helping investors recognize sentiment patterns in financial news, aiding in investment decision-making by extracting useful feature to classify the sentiment of stock market news. However, due to the vast amounts of information being put out, this can sometimes make it challenging for investors to identify what is valuable (Liang-Chih Yu 2013).

Overtime, the impact of the internet, particularly social networking, on purchasing behaviour has increased as a result of social media significantly shaping consumer preferences by influencing their attitudes and behaviours. Retailers who previously relied on physical stores have realized that social media can greatly expand their reach. Additionally, an engaging interactive presence on social can enhance brand image and provide valuable unstructured data on demand trends in a nonintrusive manner. Monitoring social media activities is an effective way to gauge customer loyalty, track sentiment towards brands or products, assess campaign impacts, and engage top influencers. Social media have become a vital platform for modern marketing strategies (Federico Neri 2012). Of the many social platforms that have been created over the years, analysing Twitter conversations has showed itself as a promising research area as these conversation provide rich, discriminative information on various topics, helping to understand people's emotions (Yili Wang 2022). Substantial research has been conducted on "Sentiment Analysis on Twitter" by various researchers. This work first focused on binary classification, which categorizes opinions or reviews into two classes: positive or negative (Vishal A. Kharde 2016). The progression of this work led to the development of a model that classifies tweets into objective, positive, and negative categories. This was done by creating a Twitter corpus by collecting tweets using the Twitter API and automatically annotating them with emoticons. Using this, sentiment classifier was developed based on the multinomial Naive Bayes method (Paroubek 2010). PoWei Liang et el. collected Twitter data using the Twitter API, focusing on three categories: camera, movie, and mobile. The data was labelled as positive, negative, or non-opinions, with opinion-containing tweets filtered out. This also utilized Naive Bayes (Po-Wei Liang 2013) Variations of Naive Bayes classifiers were also used to detect the polarity of English tweets, the baseline which classifies tweets as positive, negative, and neutral and Binary, which uses a polarity lexicon to classify tweets as positive or negative, ignoring neutral tweets (Pablo Gamallo 2014).

ML involves the scientific exploration of algorithms and statistical models that enable computer systems to execute specific tasks without explicit programming (Mahesh 2019). Algorithms that go along with this exploration are put into categories. One of which is supervised learning which is most prevalent approach in ML, which encompasses techniques such as decision trees, support vector machines, and neural networks (Liqiang Yu 2023). Neural networks are the tools used to create such machines and can be seen as a black box that produces a desired output for a given input through a process called training (Chandrahas Mishra 2017). Convolutional Neural Networks (CNNs) are frequently utilized for image classification, segmentation, object detection, video processing, natural language processing, and speech recognition. This consists of four layers: convolutional, pooling, fully connected, and non-linear layers (Purwono 2023). The application of the CNNs in relation to SA can be seen in the literature when Liao et el. employed it, to extract feature areas from global information, similar to its role in image analysis and classification. Through convolution operations, data can be processed collectively to consider relationships among features. In computer vision, CNNs extract pixel data in groups rather than individually, capturing multi-pixel information. When text data is transformed into a matrix, it can be treated similarly to an image pixel matrix, allowing the same operations to be applied for effective training of the input features in the model (Shiyang Liao 2017). Another neural network worth investigating is Long Short-Term Memory (LSTM) which is a recurrent neural network. This is used in audio and image recognition as well as natural language processing (Samsul Arifin 2023). LSTM's learning capabilities have significantly influenced multiple fields from both practical and theoretical standpoints, making it a state-of-the-art model (Greg Van Houdt 2020) .This can be seen in within the literature, particularly with respect to sentiment analysis on text reviews used by Dr. G. S. N. Murthy et el. , due to the it becoming particularly popular for sentiment classification, performing exceptionally well on a variety of problems and it being widely adopted (Dr. G. S. N. Murthy 2020). Mathieu Cliché also developed a method which combined CNN and LSTM, trained on the SemEval-2017 Twitter dataset. This method utilizes a large collection of unlabelled data and pre-trained word embeddings. This hybrid model has demonstrated significant improvements in classification accuracy. The approach involves five steps: readings the CSV file (Twitter data), pre-processing, feature extraction, and classification. In the study, the tasks are addresses using two methods: the first applies a ML-based approach to the dataset, while the second employs deep neural network-based techniques (Cliche 2017).

This research also explores the use of Python libraries, such as the Natural Language Toolkit (NLTK), specifically VADER (Valence Aware Dictionary and Sentiment Reasoner) as well as a large language model (LLM) Robustly optimized BERT (RoBERTa). When dealing with clustering outcomes from unsupervised algorithms, the VADER model is commonly used because it is well-suited for sentiment analysis in these scenarios (C.J. Hutto 2015) The VADER sentiment analysis tool uses both lexicon-based and rule-based methods to assess text, categorizing it as positive, negative, or neutral and producing a compound score that integrates these sentiments. It is especially adept at analysing social media content due to its capacity to process emojis, acronyms, slang, and contractions commonly present in these texts (Mërgim Hoti 2023). Anton Borg et el. investigated customer support data from a major Swedish telecommunication company to determine customer sentiment. The dataset included support emails from corporate clients in Sweden, organized into conversation threads, but it lacked sentiment labels. To address this, the VADER sentiment analysis framework was employed to label individual emails within these threads. The research focused on the feasibility of automatically classifying the polarity of the email content which would enable customer support teams to quickly identify and prioritize emails based on sentiment without reading entire threads. Secondly, it examined the potential for sentiment classification without detailed knowledge of grammatical rules, using a ML approach that leverages test and corresponding sentiment scores (Anton Borg 2020). Khan explains in his paper "A Large Language Model Classification Framework (LLMCF)" , that LLMs integrate the sequential text generation characteristic of autoregressive models with the transformer architecture. The self-attention mechanism within the transformer architecture allows the model to grasp relationships between words, no matter how far apart they are in the sequence. This feature overcomes the difficulties faced by earlier models in capturing long-range dependencies .This allows LLMs to produce text in a step-by-step manner while utilizing the transformer's self-attention mechanisms to understand contextual relationships (Khan 2023). Yinhan Liu et al. conducted a replication study of BERT pretraining, meticulously assessing the impact of hyperparameter adjustments and training set size. They discovered that BERT was considerably undertrained and introduced an enhanced training approach called RoBERTa. RoBERTa demonstrated that it could match or surpass the performance of subsequent BERT-based methods. Their straightforward modifications involved: (1) extending the training duration with larger batch sizes and more data; (2) eliminating the next sentence prediction task; (3) training on longer sequences; and (4) dynamically altering the masking pattern applied to the training data (Yinhan Liu 2019). The literature shows that the RoBERTa model was used in conjunction with sequence models in order to integrate the strengths of transformer models and sequence models while addressing the limitations of sequence models. The results showed that the hybrid model produced an accuracy of 96.28% on the IMDb dataset and 94.2% on the Twitter dataset (Noura A. Semary 2023).

The impact of the Coronavirus Disease 2019 (COVID-19) pandemic has profoundly disrupted daily life worldwide, affecting individuals in numerous ways (Naseem U 2021). As governments implemented lockdowns and social distancing measures, people had to adapt to new norms, from remote work and online education to changes in social interactions and healthcare access. This global crisis has not only posed significant health challenges but also led to widespread economic and social repercussions. With the influx of tweets on Twitter in regards to the pandemic (Mansoor M 2020), came the surge of SA used to investigate public sentiment during this period. Jalil et el. investigate these very sentiments by applying ML and deep learning (DL)-based classification methods, focusing on tweets related to the coronavirus in their article "COVID-19 Related Sentiment Analysis Using State-of-the-Art Machine Learning and Deep Learning Techniques". Their study introduces a Transformation-based Multi-depth DistilBERT model designed to identify sentiments within these tweets, allowing for the automatic extraction of sentiment-related information without human intervention. The research also includes a comprehensive comparison of existing ML and DL text classification methods, demonstrating that their proposed model outperforms previous methods on real-world datasets (Z. Jalil 2022). Another landscape that has been worth investigating is the application of how SA can influence public policy and economics. Understanding public sentiment is vital for assessing the services offered by public administration and for creating new processes. This analysis aims to aid decision-makers in making informed decisions that foster growth and innovation, thereby improving the everyday lives of the community (A. Corallo 2015). There are studies that tackle the breakdown of food imports as seen in Kaitibie et el. Paper on the "Analysis of Food Imports in a Highly Import Dependent Economy", that employ enhanced panel data methods, to demonstrate that several factors significantly influence Qatar's food imports (Simeon Kaitibie 2017) as well as studies that examine the geographic patterns of sentiment in tweets related to food security among Malaysians, identifying areas with predominantly positive, negative, or neutral sentiment. Rusli et el. Research gave insight into public sentiment regarding increasing food prices, as expressed on social media, as being instrumental for policymakers and stakeholder in addressing these issues (N Rusli 2024) However, there is not a lot of research that combines or encapsulates the use of SA specifically in relation to imported food items especially within the Caribbean context in Trinidad and Tobago. This gap in the research has inspired the use of SA to show that it can be used to make informed decisions related to food imports by providing insights into public opinion on imported food items which can in turn help policymakers in Trinidad and Tobago make data-driven decisions to manage the import bill. The application of the models proposed in the methodology aims to address the need for more robust models for social media AS and the impact of specific events like COVID-19 on sentiment trends.

**Section 3: Methodology**

In order to analyse Twitter data regarding specific imported food items in Trinidad and Tobago using various ML algorithms, an investigation into whether or not these algorithms can effectively and accurately identify sentiment will be carried out. Natural language processing (NLP) methods are integral to this research for analysing sentiment in Twitter data related to imported food items in Trinidad and Tobago. The methodology comprises three primary sections: data collection, data pre-processing and experiments.

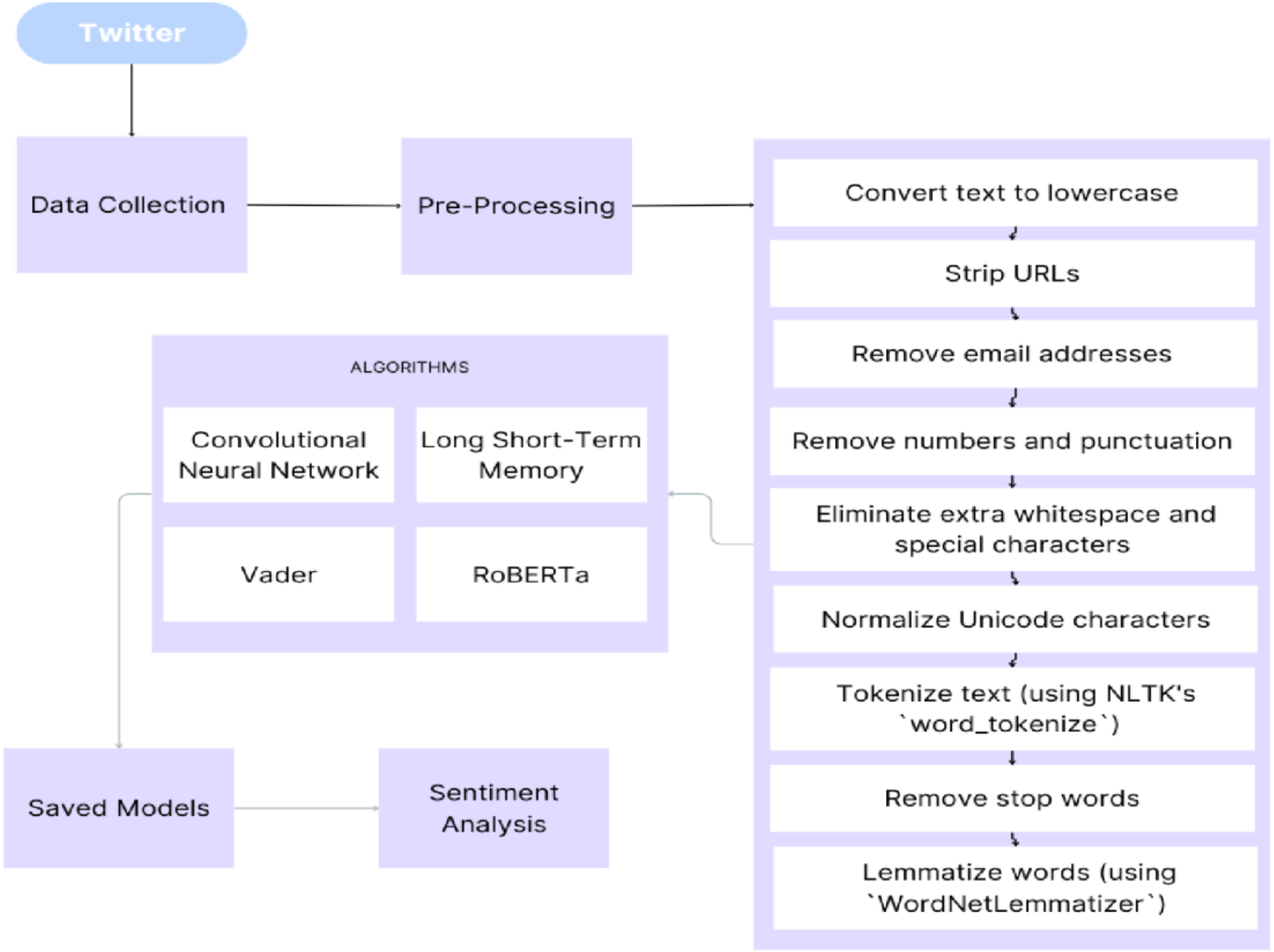

*Figure 1. Flowchart of Sentiment Classification of Twitter data*

The process begins with data collection, where Twitter data is gathered using a web scraping tool. This tool will be set up to extract tweets related to imported food items in Trinidad and Tobago. Once the data is collected, it undergoes a series of pre-processing steps. In the experiments phase, different ML algorithms are applied to the pre-processed data. Using Python, these models will be trained and evaluated to determine their accuracy and effectiveness in sentiment analysis. Various configurations and hyperparameters will be tested to identify the optimal settings for each model. The final section involves applying the optimized models to sentiment analysis. The models from the experiments will be used to analyse a use case scenario to determine whether the models performance can be used to respond to public opinion on specific imported food items. These steps will be further expanded in the subsections mentioned.

3.1 Data Collection

By employing Twitter's advanced search feature using a desktop browser, the data collection process began by gathering tweets containing specific keywords. The keywords included in this search were apples, grapes, lobster, pears, strawberries, and peaches, which are six food items that are imported by Trinidad and Tobago. The rationale behind selecting these specific keywords stems from an article where Trade and Industry Minister Paula Gopee-Scoon highlighted that Trinidad and Tobago spent approximately $78 million a year over the last three years on the importation of luxury food items, including smoked salmon, lobster, grapes, apples, pears, strawberries, and peaches (TACK 2021). This underscores the significance of these items in the local import landscape and the need to understand public sentiment towards them. The search was further refined by setting the location to 'near me' to ensure that only the tweets that were

geographically relevant and the language filter was set to English. The date filter was not set to any specific range, which resulted in a dataset that ranged from 2018 to 2024. This broad date range was chosen because the tweets pertaining to the imported food items were limited, and a wider range allowed for the collection of as many relevant tweets as possible. Using a custom web scraping tool that utilized the BeautifulSoup library, detailed information was extracted from the tweets. This was possible due to the BeautifulSoup library being able to extract the relevant tweet information from the HTML structure of each tweet's page, extracting attributes such as the tweet text, timestamp, view count, location, username, and engagement metrics such as replies, retweets, likes, and bookmarks (see Appendix 1 for the code). In total, 207 tweets were obtained, and each tweet's sentiment was manually labelled.

## 3.2 Data Pre-processing

Once the tweet data was collected, several pre-processing steps were implemented to prepare the data for ML and natural language processing tasks. Each step in the pre-processing pipeline was carefully chosen to address specific challenges in working with textual data from social media, ensuring that the resulting dataset is clean, consistent, and suitable for sentiment analysis. The detailed pre-processing steps and the rationale behind each are as follows:

- Converting Text to Lowercase: All text was converted to lowercase to ensure uniformity and reduce variability caused by capitalization. For instance, "Apple" and "apple" would be considered the same word, which helps in maintaining consistency across the dataset and avoids treating them as different words.
- Removing URLs: URLs were stripped from the text to eliminate unnecessary noise. URLs do not contribute to the sentiment of a tweet and could introduce irrelevant information, thereby affecting the performance of the sentiment analysis models.
- Removing Email Addresses: Email addresses were removed using regular expressions to clean the text and avoid irrelevant information which can introduce noise into the dataset.
- Removing Numbers and Punctuation: Numbers and punctuation were removed to focus on the textual content. This step helps in simplifying the text and reducing complexity, which is crucial for the performance of natural language processing models.
- Removing Extra Whitespace and Special Characters: Extra whitespace and special characters were removed to clean the text. This step ensures that the text is neatly formatted and free of unnecessary spaces and characters that could interfere with tokenization and subsequent analysis.
- Normalizing Unicode Characters: Unicode characters were normalized to standardize the text format and remove emojis. This step is important for consistency and helps in avoiding issues with different character encodings that can complicate text processing.
- Tokenization: The text was broken down into individual words using the 'word_tokenize' function from the NLTK library. Tokenization is the process of splitting text into individual tokens (words), which is essential for further analysis like lemmatization and stop-word removal.
- Removing Stop Words: Stop words (common words like "and", "the", "is") were removed to eliminate words that do not add significant meaning. This step helps in focusing on the important words that carry sentiment and meaning, thereby improving the effectiveness of the sentiment analysis models.
- Lemmatization: Words were reduced to their base form using the 'WordNetLemmatizer'. Lemmatization helps in normalizing words to their root form (e.g., "eating" to "eat"), which reduces the dimensionality of the dataset and ensures that different forms of a word are treated as a single item.

After these processes, three datasets were obtained and saved as CSV files:

1. Imbalanced Dataset: This dataset includes the clean tweets spanning from 2018 to 2024. It was derived from the initial web scraping and preprocessing steps described above. This dataset provides a comprehensive view of tweets over a broad timeframe, capturing a wide range of sentiments and opinions.
2. Balanced Dataset: Derived from the initial dataset, this dataset was created by applying under-sampling techniques to balance the class distribution. It contains 120 tweets, ensuring an equal representation of each class (positive, negative, neutral) to prevent model bias and improve performance. Balancing the dataset is crucial for training models that are not skewed towards the majority class, thus enhancing the reliability of sentiment predictions.
3. Temporal Subsets: The final dataset was derived from the initial dataset by sorting tweets by year to create two subsets: pre-COVID (tweets from 20182020) and post-COVID (tweets from 2020-2024). This temporal division allows for the analysis of sentiment trends over time, particularly assessing the impact of the COVID-19 pandemic on public sentiment towards imported food items.

These pre-processing steps ensure that the text data is in a suitable format for ML algorithms, enhancing the quality and performance of sentiment analysis models applied in subsequent experiments.

## 3.3 Experimental Design

To evaluate the performance of sentiment analysis, ten experiments were conducted using four different algorithms: CNN, LSTM, VADER, and RoBERTa. The dataset was manually labelled into positive, negative, and neutral sentiments. Examples of tweets from our dataset that were classified can be found in Appendix 2. All experiments were trained, validated, and tested using an 80/10/10 split and a consistent random seed for reproducibility. A SA process for the VADER sentiment analyser and defining polarity weights for the RoBERTa model will be initialized. It includes a function to calculate the RoBERTa polarity score by converting sentiment scores into probabilities, scaling these probabilities, and returning a final polarity value. The script applies the VADER analyser to both training and testing datasets, calculating sentiment scores for cleaned text and classifying these scores into positive, neutral, or negative sentiments based on the composite score. The classification rules are:

- Positive sentiment: compound score >= 0.05
- Neutral sentiment: -0.05 < compound score < 0.05
- Negative sentiment: compound score <= -0.05

Similarly, a RoBERTa sentiment analysis pipeline is initialized and applied to the datasets, transforming the text into sentiment scores, which are then used to determine polarity and sentiment classification using the same thresholds. Finally, the performance of both VADER and RoBERTa models is evaluated and visualized using the defined functions, enabling a comparison of their effectiveness in sentiment analysis.

The experiments are described as follows:

### 3.3.1 Convolutional Neural Network (CNN) Experiments

Experiment 1.

The first experiment evaluated a CNN model using fixed hyperparameters. The model architecture included an embedding layer, a convolutional layer, a global max pooling layer, dense layers, and dropout for regularization. The text data was tokenized and padded to a maximum sequence length, and labels were one-hot encoded. The model was compiled with categorical cross-entropy loss and the Adam optimizer, and trained for 10 epochs with a batch size of 16. A checkpoint call-back saved the best model weights based on validation loss, and the model's performance was evaluated on the test set.

Experiment 2.

The second experiment focused on optimizing the CNN model by varying hyperparameters: activation functions (ReLU, tanh, sigmoid), optimizers (Adam, RMSprop), batch sizes (4, 16, 32, 64), and epochs (5, 10, 15). Each configuration was assessed based on validation accuracy to identify the best-performing model, which was then evaluated on the test set.

Experiment 3.

The third experiment addressed class imbalance by conducting hyperparameter optimization on a balanced dataset containing 120 tweets. The same range of hyperparameters as in Experiment 2 was tested. The best model was selected based on validation accuracy and evaluated on the test set to assess its generalization capability.

### 3.3.2 Long Short-Term Memory (LSTM) Experiments

Experiment 4.

The fourth experiment evaluated an LSTM model with fixed hyperparameters on the pre-processed dataset. The model architecture included an embedding layer, an LSTM layer, dense layers, and dropout for regularization. The model was trained for 10 epochs with a batch size of 16, using the Adam optimizer. The model's performance was then evaluated on the test set.

Experiment 5.

The fifth experiment optimized the LSTM model by varying hyperparameters: the number of LSTM units (64, 128), activation functions (ReLU, tanh), optimizers (Adam, RMSprop, SGD, Adadelta), batch sizes (4, 16, 32), and epochs (5, 10, 15, 20). The best model configuration was selected based on validation accuracy and evaluated on the test set.

Experiment 6.

The sixth experiment addressed class imbalance by conducting hyperparameter optimization on the balanced dataset. The same range of hyperparameters as in Experiment 5 was tested. The best model was selected based on validation accuracy and evaluated on the test set.

### 3.3.3 VADER and RoBERTa Experiments

Experiment 7.

This experiment analysed the sentiment of tweets using VADER and RoBERTa models on the original dataset. The performance of both models was evaluated using accuracy, F1-score, precision, and recall to classify tweets into positive, negative, and neutral sentiments.

Experiment 8.

In this experiment, VADER and RoBERTa models were used on a balanced dataset. The performance metrics were the same as in Experiment 7, and the results were compared to assess the impact of class balancing.

Experiment 9.

The ninth experiment focused on binary sentiment classification by analysing only positive and negative sentiments, excluding neutral sentiments from the original dataset. The performance of VADER and RoBERTa models was evaluated using binary classification metrics (accuracy, F1-score, precision, and recall).

### 3.3.4 Use Case Experiment

Experiment 10.

The tenth experiment compared trends in public sentiment pre- and post-COVID19. The dataset was split into two subsets: pre-COVID (tweets before March 2020) and post-COVID (tweets from March 2020 onwards). The best-performing models from previous experiments were used and their performances were observed on these datasets. The best of those models was used to classify tweets into sentiment categories, and the results from both periods were compared to analyse differences in sentiment trends.

The experiments were designed to not only evaluate individual model performances but also to provide a comparative analysis of different ML algorithms. The rationale behind selecting specific hyperparameters for variation in each experiment was to explore their impact on model performance and identify the most effective configurations. For evaluation, while metrics like accuracy, F1-score, precision, and recall were used, the primary metric prioritized for model evaluation was validation accuracy. This metric was chosen because it provides a direct measure of how well the model generalizes to unseen data during training. Additionally, a comparative analysis was conducted to assess the performance of CNN, LSTM, VADER, and RoBERTa in terms of accuracy and efficiency for sentiment analysis. This comprehensive approach aimed to identify the strengths and weaknesses of each algorithm, providing a understanding of their capabilities in different contexts and highlighting the most effective models for sentiment analysis of tweets related to imported food items in Trinidad and Tobago.

3.3.5 Sentiment Analysis

The sentiment analysis in this study utilized the best-performing models identified from our extensive experimentation process. These models were applied to the dataset to classify tweets into positive, negative, and neutral sentiments. The analysis aimed to uncover the public's sentiment towards specific imported food items and assess how these sentiments might influence import decisions. Furthermore, a comparative analysis of sentiment trends pre- and post-COVID-19 was done. The dataset was divided based on tweet dates to create pre- and postCOVID subsets. Sentiment analysis models were applied separately to each subset, and the resulting sentiment trends were analysed to understand the pandemic's impact on public opinion. To ensure robust sentiment analysis, a range of ML models, including CNN, LSTM, VADER, and RoBERTa, were employed and optimized their performance through hyperparameter tuning. The final models were evaluated using metrics such as accuracy, F1-score, precision, and recall, ensuring reliable sentiment classification. By leveraging these advanced models and techniques, the methodology provides a comprehensive approach to analysing Twitter data for sentiment analysis. This approach addresses key research questions related to public sentiment towards imported food items in Trinidad and Tobago and explores how these sentiments have evolved in response to the COVID-19 pandemic. The findings from this analysis offer valuable insights into the public's perception and its potential implications for import policies and decisions.

**Section 4: Detailed Results And Analysis**

The results of this study aimed to evaluate the performance of various ML algorithms CNN, LSTM, VADER, and RoBERTa for sentiment analysis of Twitter data related to imported food items in Trinidad and Tobago. The experiments focused on analysing the accuracy and efficiency of these models under different configurations and conditions, including fixed and varying hyperparameters, imbalanced and balanced datasets, and multi-class versus binary sentiment classifications.

*Table 1: CNN Model Performance*

| Experiment No. | Dataset | Hyperparameters | Accuracy Score | F1 Score | Test Loss |
|---|---|---|---|---|---|
| 1 (Best) | Imbalanced | Fixed | 0.5714 | 0.4908 | 0.9175 |
| 2 | Imbalanced | Varying | 0.5238 | 0.4359 | 0.9685 |
| 3 | Balanced | Varying | 0.4167 | 0.3549 | 1.0966 |

*Table 2: LSTM Model Performance*

| Experiment No. | Dataset | Hyperparameters | Accuracy Score | F1 Score | Test Loss |
|---|---|---|---|---|---|
| 4 | Imbalanced | Fixed | 0.3810 | 0.2102 | 1.1250 |
| 5 (Best) | Imbalanced | Varying | 0.3810 | 0.2102 | 1.1060 |
| 6 | Balanced | Varying | 0.2500 | 0.1000 | 1.1137 |

*Table 3: VADER Model Performance*

| Experiment No. | Dataset | Sentiment Classification | Accuracy Score | F1 Score |
|---|---|---|---|---|
| 7 | Imbalanced | Multi-Class | 0.6190 | 0.6192 |
| 8 | Balanced | Multi-Class | 0.5278 | 0.5049 |
| 9 (Best) | Imbalanced | Binary | 0.6977 | 0.6935 |

*Table 4: RoBERTa Model Performance*

| Experiment No. | Dataset | Sentiment Classification | Accuracy Score | F1 Score |
|---|---|---|---|---|
| 7 (Best) | Imbalanced | Multi-Class | 0.4921 | 0.3246 |
| 8 | Balanced | Multi-Class | 0.3333 | 0.1667 |
| 9 | Imbalanced | Binary | 0.2791 | 0.1218 |

*Table 5: Sentiment Trends Pre- and Post-COVID-19*

| Period | Model | Dataset | Accuracy Score | F1 Score | Test Loss |
|---|---|---|---|---|---|
| Pre-COVID | CNN | Imbalanced | 0.6667 | 0.5333 | 1.0913 |
| Post-COVID | CNN | Imbalanced | 0.3333 | 0.1667 | 1.0501 |
| Pre-COVID | LSTM | Imbalanced | 0.4500 | 0.2793 | 1.0266 |
| Post-COVID | LSTM | Imbalanced | 0.4500 | 0.2793 | 1.0465 |
| Pre-COVID | VADER | Binary | 0.8333 | 0.8148 | - |
| Post-COVID | VADER | Binary | 0.7500 | 0.7500 | - |
| Pre-COVID | RoBERTa | Multi-Class | 0.4444 | 0.2735 | - |
| Post-COVID | RoBERTa | Multi-Class | 0.5000 | 0.3333 | - |

*Table 6: Hyperparameter Settings for Best Models*

| Model | Hyperparameters | Best Setting |
|---|---|---|
| CNN | Activation Function | ReLU |
| | Optimizer | Adam |
| | Batch Size | 16 |
| | Epochs | 10 |
| | Dropout Rate | 0.5 |
| LSTM | Number of LSTM Units | 64 |
| | Activation Function | ReLU |
| | Optimizer | Adam |
| | Batch Size | 16 |
| | Epochs | 10 |
| | Dropout Rate | 0.5 |
| VADER | Threshold for Sentiment Classification | 0.05 |
| RoBERTa | Pre-trained Model | base |
| | Fine-tuning Epochs 3 | 3 |
| | Learning Rate | 2e-5 |

4.1 CNN Model Performance

*Table 7: Performance Metrics of CNN Models on the Imbalanced Dataset with Fixed and Varying Hyperparameters*

| Model | Hyperparameters | Accuracy Score | F1 Score | Test Loss |
|---|---|---|---|---|
| CNN | Fixed | 0.5714 | 0.4908 | 0.9175 |
| CNN | Varying | 0.5238 | 0.4359 | 0.9685 |

The CNN model with fixed hyperparameters performs better than the one with varying hyperparameters. This suggests that the initial fixed configuration was more effective for the imbalanced dataset. Despite attempts to optimize, the varying hyperparameters resulted in a slight decrease in both accuracy and F1 score.

*Table 8: Performance Metrics of LSTM Models on the Imbalanced Dataset with Fixed and Varying Hyperparameters*

| Model | Accuracy Score | F1 Score | Test Loss |
|---|---|---|---|
| CNN | 0.4167 | 0.3549 | 1.0966 |

On the balanced dataset, the CNN model shows a decrease in performance compared to the imbalanced dataset. The lower accuracy and F1 score indicate that balancing the dataset did not improve the model's ability to generalize, possibly due to overfitting on the smaller balanced dataset.

4.2 LSTM Model Performance

*Table 9: Performance Metrics of LSTM Models on the Imbalanced Dataset with Fixed and Varying Hyperparameters*

| Model | Hyperparameters | Accuracy Score | F1 Score | Test Loss |
|---|---|---|---|---|
| LSTM | Fixed | 0.3810 | 0.2102 | 1.1250 |
| LSTM | Varying | 0.3810 | 0.2102 | 1.1060 |

The LSTM model exhibits identical performance with both fixed and varying hyperparameters on the imbalanced dataset. This indicates that the variations in hyperparameters did not have any impact on the model's performance, suggesting a need for more sophisticated tuning or a different architecture for better results.

*Table 10: Performance Metrics of LSTM Models on the Balanced Dataset with Varying Hyperparameters*

| Model | Accuracy Score | F1 Score | Test Loss |
|---|---|---|---|
| LSTM | 0.2500 | 0.1000 | 1.1137 |

On the balanced dataset, the LSTM model performs worse than on the imbalanced dataset. The accuracy and F1 score are significantly lower, indicating that the LSTM model struggles to learn effectively from the balanced dataset, possibly due to the reduced size or complexity of the balanced data.

4.3 VADER and RoBERTa Model Performance

*Table 11: Performance Metrics of VADER and RoBERTa Models for Multi-class Sentiment Analysis on the Imbalanced Dataset.*

| Model | Sentiment Classification | Accuracy Score | F1 Score |
|---|---|---|---|
| VADER | Multi-Class | 0.6190 | 0.6192 |
| RoBERTa | Multi-Class | 0.4921 | 0.3246 |

VADER outperforms RoBERTa in multi-class sentiment analysis on the imbalanced dataset. VADER shows a higher accuracy and F1 score, indicating better performance in classifying the sentiments. RoBERTa struggles, possibly due to its architecture being less suited for this specific task without fine-tuning.

*Table 12: Performance Metrics of VADER and RoBERTa Models for Multi-class Sentiment Analysis on the Balanced Dataset*

| Model | Sentiment Classification | Accuracy Score | F1 Score |
|---|---|---|---|
| VADER | Multi-Class | 0.5278 | 0.5049 |
| RoBERTa | Multi-Class | 0.3333 | 0.1667 |

Both models see a drop in performance on the balanced dataset compared to the imbalanced dataset. However, VADER still outperforms RoBERTa, indicating it is more robust for multi-class sentiment analysis. RoBERTa's performance further declines, suggesting that it does not handle the balanced dataset well without further tuning.

*Table 13: : Performance Metrics of VADER and RoBERTa Models for Binary Sentiment Analysis on the Imbalanced Dataset*

| Model | Sentiment Classification | Accuracy Score | F1 Score |
|---|---|---|---|
| VADER | Binary | 0.6977 | 0.6935 |
| RoBERTa | Binary | 0.2791 | 0.1218 |

VADER significantly outperforms RoBERTa in binary sentiment analysis on the imbalanced dataset. The higher accuracy and F1 score for VADER suggest that it is much more effective in distinguishing between positive and negative sentiments. RoBERTa's performance is poor, indicating substantial difficulty in binary classification in this context.

*Table 14: Performance Metrics of VADER and RoBERTa Models for Binary Sentiment Analysis on the Balanced Dataset*

| Model | Sentiment Classification | Accuracy Score | F1 Score |
|---|---|---|---|
| VADER | Binary | 0.5833 | 0.5313 |
| RoBERTa | Binary | 0.5000 | 0.3333 |

On the balanced dataset, VADER again outperforms RoBERTa, though both models show a decrease in performance compared to the imbalanced dataset. VADER's accuracy and F1 score are still higher, indicating its better generalization ability. RoBERTa's performance improves slightly compared to its multi-class results, but it remains less effective than VADER for binary sentiment classification.

4.4 Sentiment Trends Pre- and Post-COVID-19

*Table 15: Performance Metrics of CNN Models on Pre- and Post-COVID-19 Datasets*

| Period | Model | Dataset | Accuracy Score | F1 Score | Test Loss |
|---|---|---|---|---|---|
| Pre-COVID | CNN | Imbalanced | 0.6667 | 0.5333 | 1.0913 |
| Post-COVID | CNN | Imbalanced | 0.3333 | 0.1667 | 1.0501 |

The CNN model shows a significant drop in performance post-COVID-19. The accuracy and F1 score are considerably lower in the post-COVID-19 dataset, indicating that the sentiment distribution or characteristics of the tweets might have changed substantially after the onset of the pandemic, affecting the model's performance.

*Table 16: Performance Metrics of LSTM Models on Pre- and Post-COVID-19 Datasets*

| Period | Model | Dataset | Accuracy Score | F1 Score | Test Loss |
|---|---|---|---|---|---|
| Pre-COVID | LSTM | Imbalanced | 0.4500 | 0.2793 | 1.0266 |
| Post-COVID | LSTM | Imbalanced | 0.4500 | 0.2793 | 1.0465 |

The LSTM model shows no change in accuracy and F1 score pre- and post-COVID19, though there is a slight increase in test loss post-COVID-19. This suggests that the LSTM model's performance is consistent over time but does not improve with the new data characteristics post-pandemic.

*Table 17: Performance Metrics of VADER Models on Pre- and Post-COVID-19 Datasets*

| Period | Model | Dataset | Accuracy Score | F1 Score | Test Loss |
|---|---|---|---|---|---|
| Pre-COVID | VADER | Binary | 0.8333 | 0.8148 | - |
| Post-COVID | VADER | Binary | 0.4444 | 0.2735 | - |

VADER shows a significant decrease in performance post-COVID-19. The accuracy and F1 score drop sharply, indicating that the binary sentiment classification became more challenging after the onset of the pandemic, likely due to shifts in sentiment expression or topic distribution.

*Table 18: Performance Metrics of RoBERTa Models on Pre- and Post-COVID-19 Datasets*

| Period | Model | Dataset | Accuracy Score | F1 Score | Test Loss |
|---|---|---|---|---|---|
| Pre-COVID | RoBERTa | Multi-Class | 0.7500 | 0.7500 | - |
| Post-COVID | RoBERTa | Multi-Class | 0.5000 | 0.3333 | - |

RoBERTa's performance also declines post-COVID-19. The drop in both accuracy and F1 score suggests that the model was less effective at classifying multi-class sentiments after the pandemic began, pointing to possible changes in the language or sentiment patterns in the tweets.

4.5 Figures and Additional Analysis

4.5.1 CNN Model Analysis

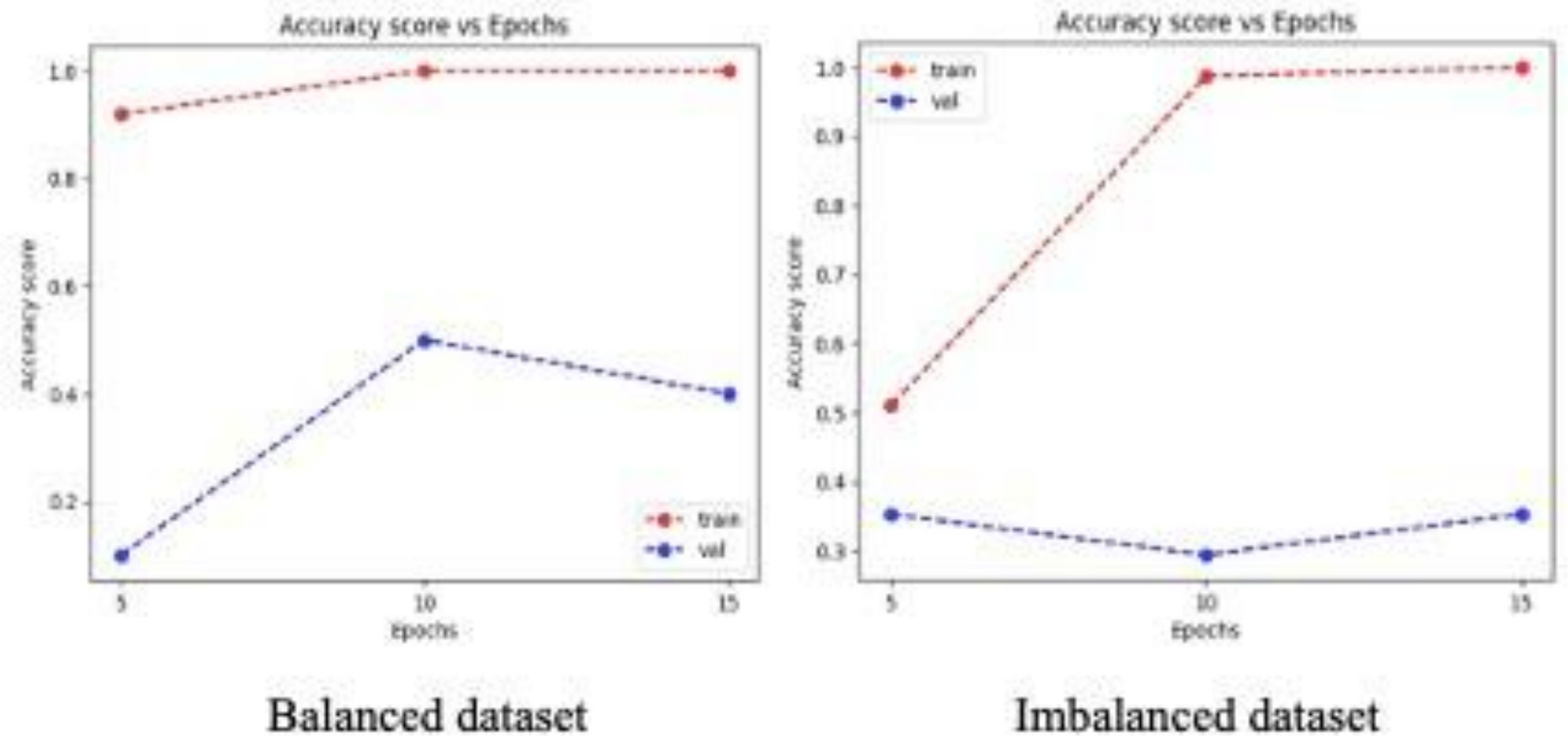

*Figure 2: CNN - Accuracy vs. Epochs (Balanced vs. Imbalanced Dataset)*

In figure 2, the balanced dataset shows that the training accuracy starts high and slightly increases as the epochs progress. This indicates that the model is learning well from the training data. The validation accuracy is lower than the training accuracy and shows an increasing trend initially, but then decreases after 10 epochs. This suggests that the model starts overfitting after a certain number of epochs. Conversely, for the imbalanced dataset, the training accuracy increases steadily with the number of epochs. This shows the model is improving its learning with more epochs. The validation accuracy also shows an initial increase but then decreases, indicating potential overfitting as the model continues to train beyond a certain number of epochs.

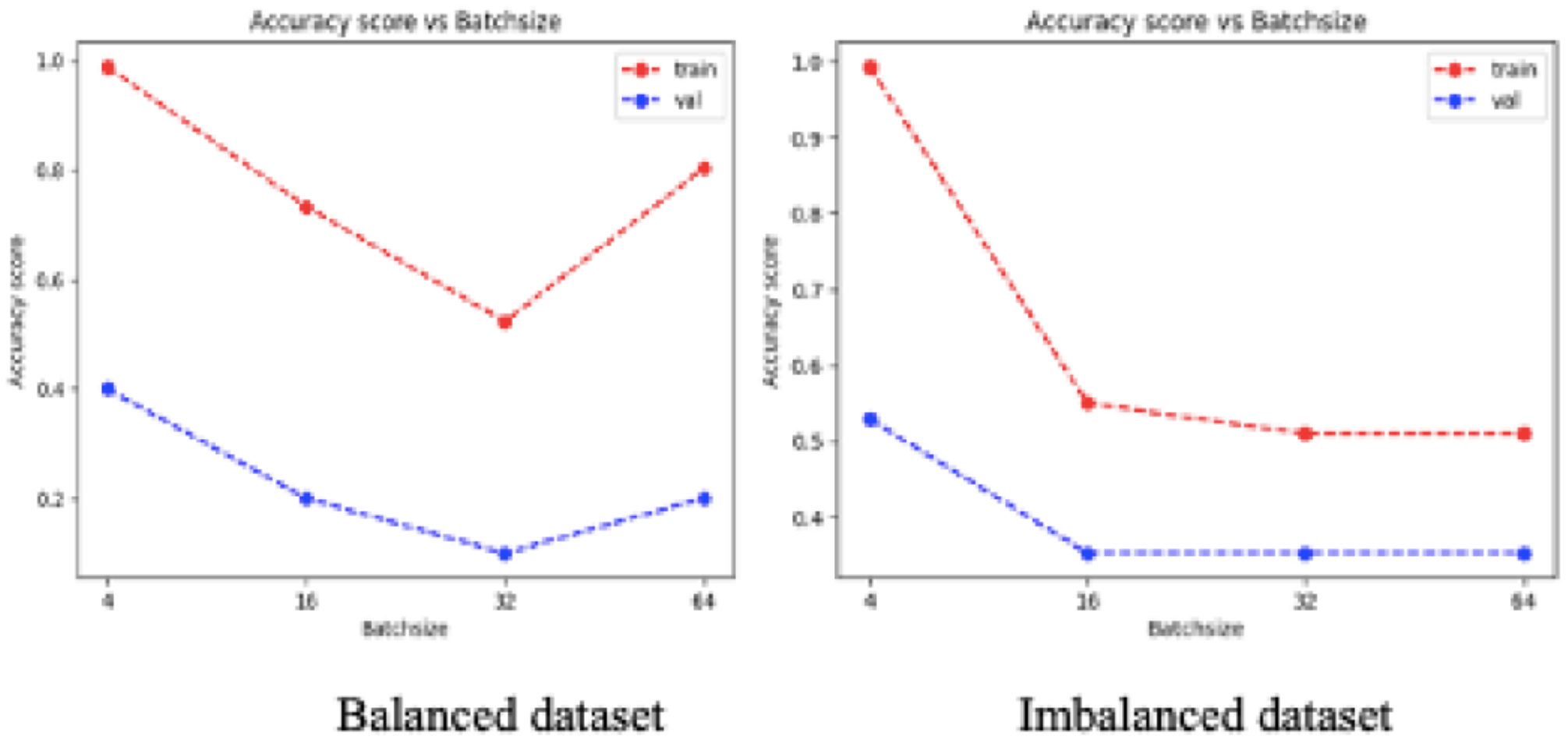

*Figure 3: CNN - Accuracy vs. Batch Size (Balanced vs. Imbalanced Dataset)*

For the balanced dataset, the training accuracy decreases as the batch size increases. This could indicate that smaller batch sizes allow for more frequent updates and better learning. The validation accuracy also decreases with increasing batch size, although there is a slight improvement at a batch size of 64. Generally, the model performs better with smaller batch sizes. Similarly, for the imbalanced dataset, the training accuracy decreases with increasing batch size. Smaller batch sizes seem to benefit the training process. The validation accuracy decreases as the batch size increases, which aligns with the trend seen in the balanced dataset. Again, the model tends to perform better with smaller batch sizes, but there's a slight increase at a batch size of 32.

4.5.2 LSTM Model Analysis

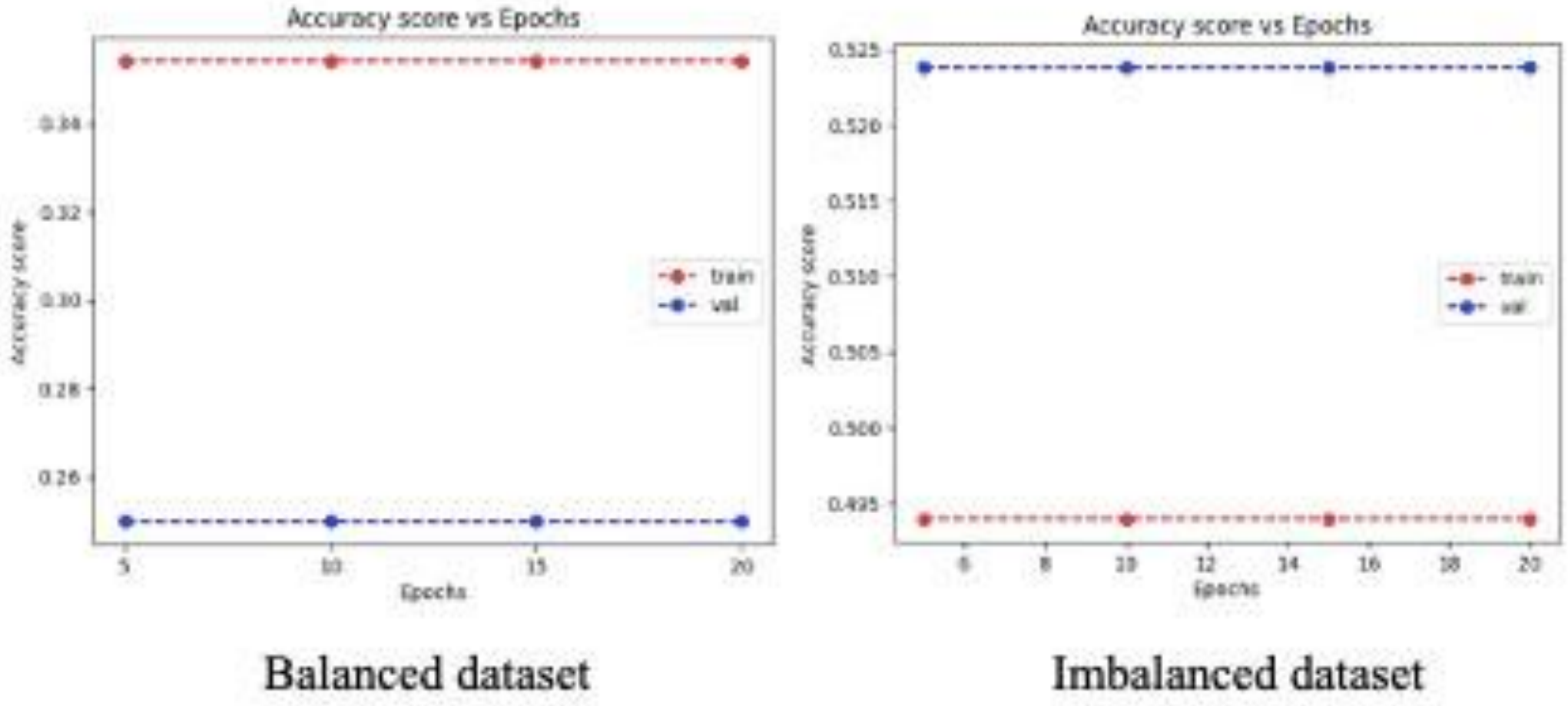

*Figure 4: LSTM - Accuracy vs. Epoch (Balanced vs. Imbalanced Dataset)*

In figure 4, it is observed that the training accuracy remains constant and low across different batch sizes for both the balanced and imbalanced datasets. This indicates that changing the batch size does not impact the learning of the LSTM model on the imbalanced dataset. Similarly, the validation accuracy also stays constant and low for both datasets.

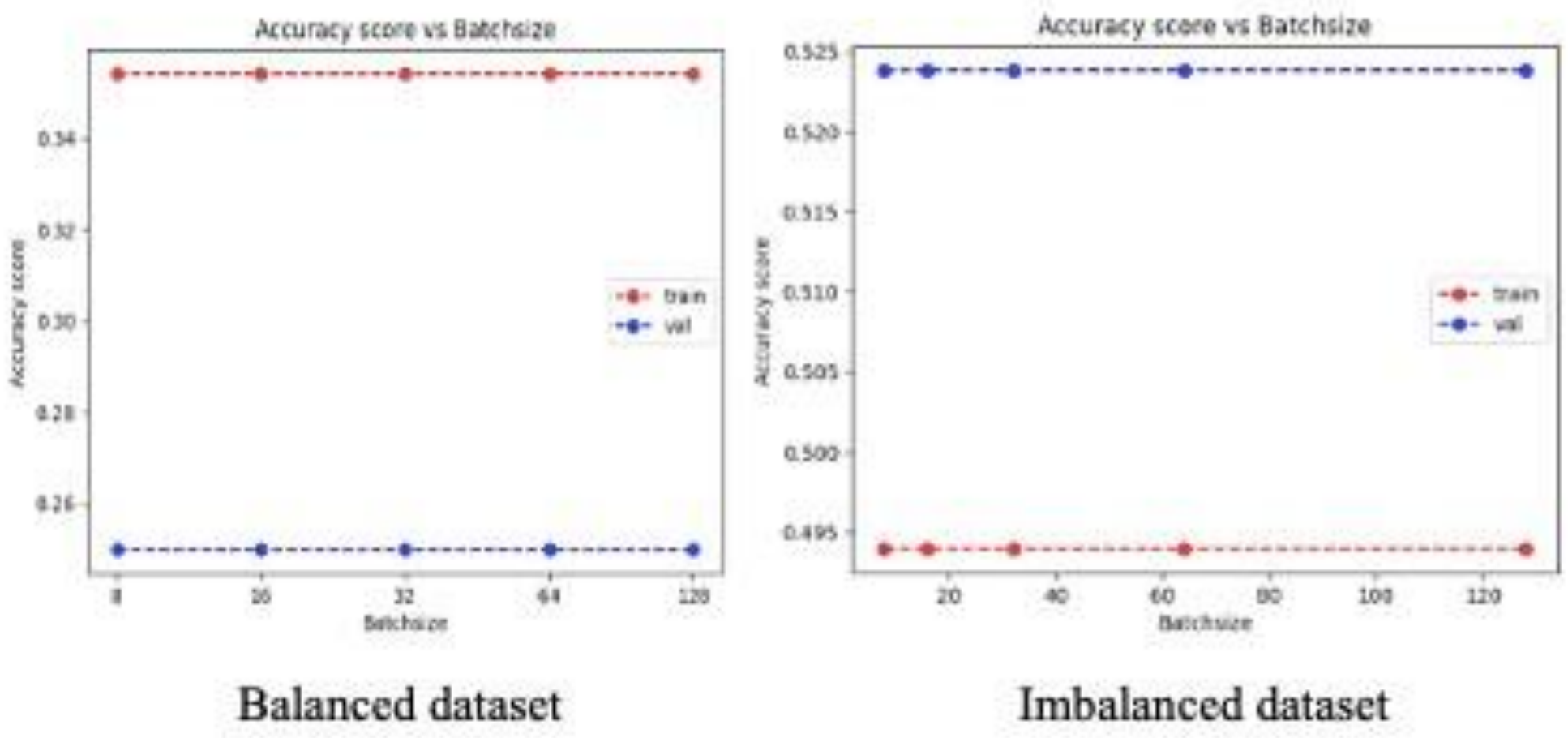

*Figure 5: LSTM - Accuracy vs. Batch Size (Balanced vs. Imbalanced Dataset)*

Similar observations from figure 4 apply to figure 5, where the training accuracy remains constant and low across different batch sizes for both balanced and imbalanced datasets, and the validation accuracy also stays constant and low for both datasets.

4.5.3 VADER and RoBERTa Model Analysis

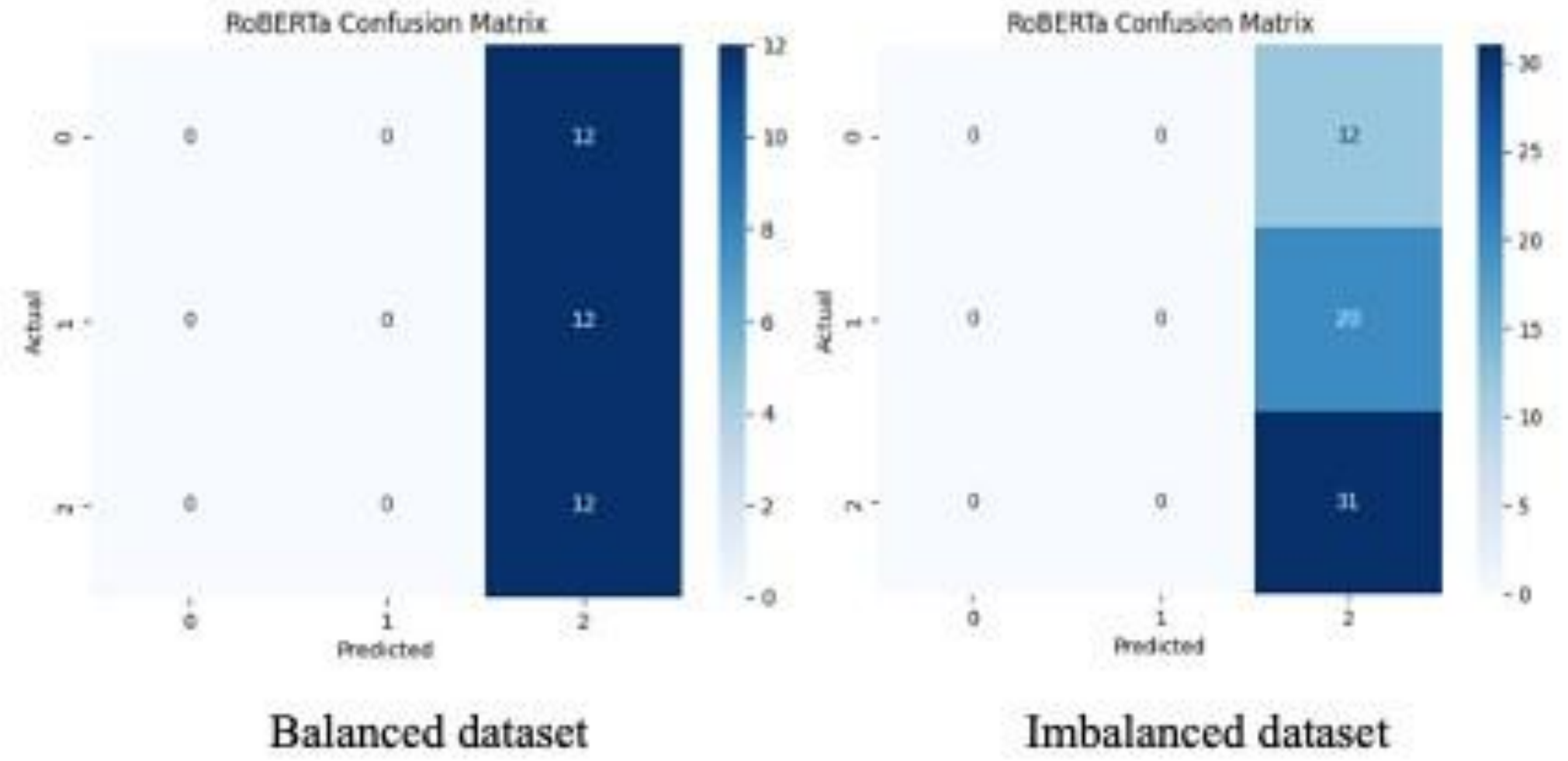

*Figure 6: RoBERTa - Confusion Matrix for Multi-class Sentiment Analysis (Balanced vs. Imbalanced Dataset)*

The RoBERTa model shows a significant bias towards predicting the positive class (Class 2), failing to correctly classify any instances as negative (Class 0) or neutral (Class 1) on the imbalanced dataset. The precision and recall for Classes 0 and 1 are both 0.00, resulting in an overall low F1-score and accuracy of 0.49. The model performs slightly better on the balanced dataset but still shows a significant bias towards the positive class (Class 2), failing to correctly classify any instances as negative (Class 0) or neutral (Class 1). The precision and recall for Classes 0 and 1 remain 0.00, with an overall accuracy of 0.33.

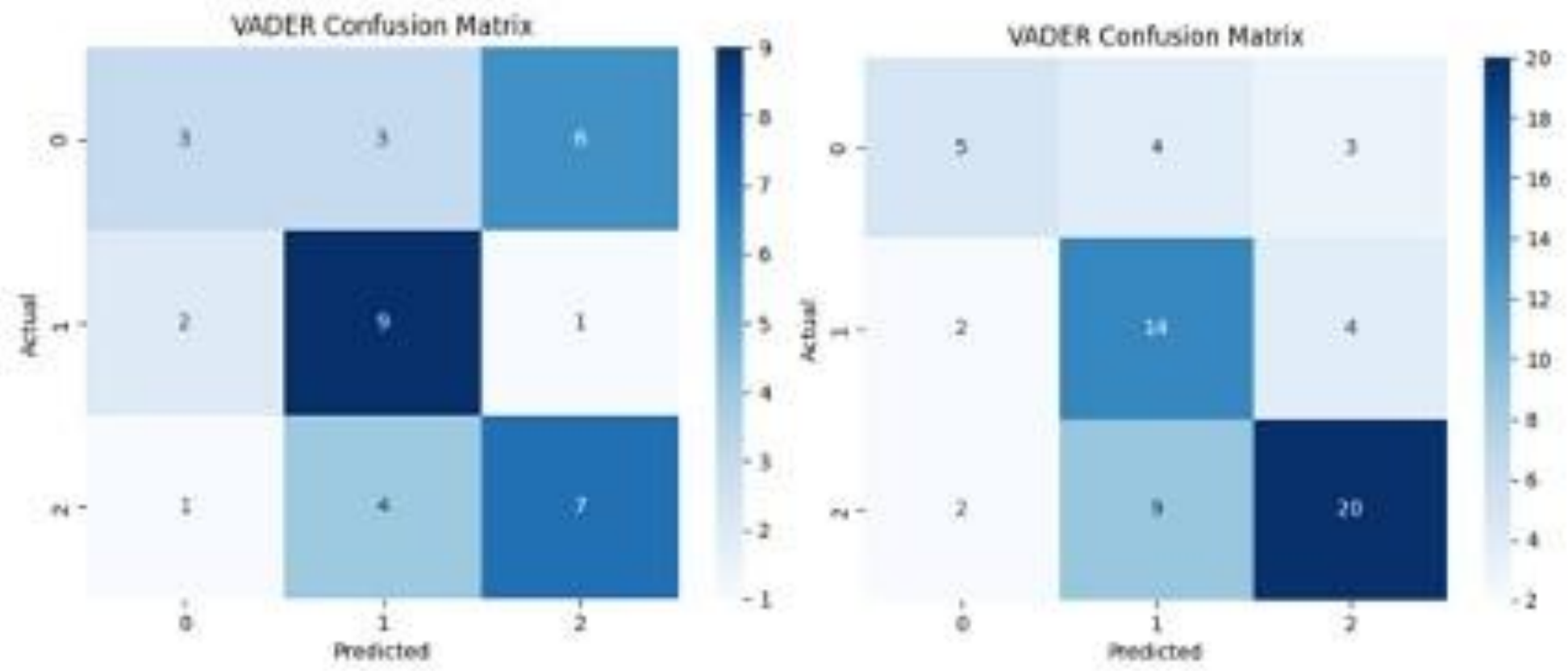

Balanced dataset Imbalanced dataset

*Figure 7: VADER - Confusion Matrix for Multi-class Sentiment Analysis (Balanced vs. Imbalanced Dataset)*

The VADER model shows a more distributed classification but still struggles with correctly classifying all instances in the imbalanced dataset. It performs relatively better in identifying positive instances (Class 2) with higher precision, recall, and F1-score, resulting in an overall accuracy of 0.62. The model shows improved performance on the balanced dataset, with higher precision and recall for both neutral (Class 1) and positive (Class 2) classes. However, there are still notable misclassifications, resulting in an overall accuracy of 0.53.

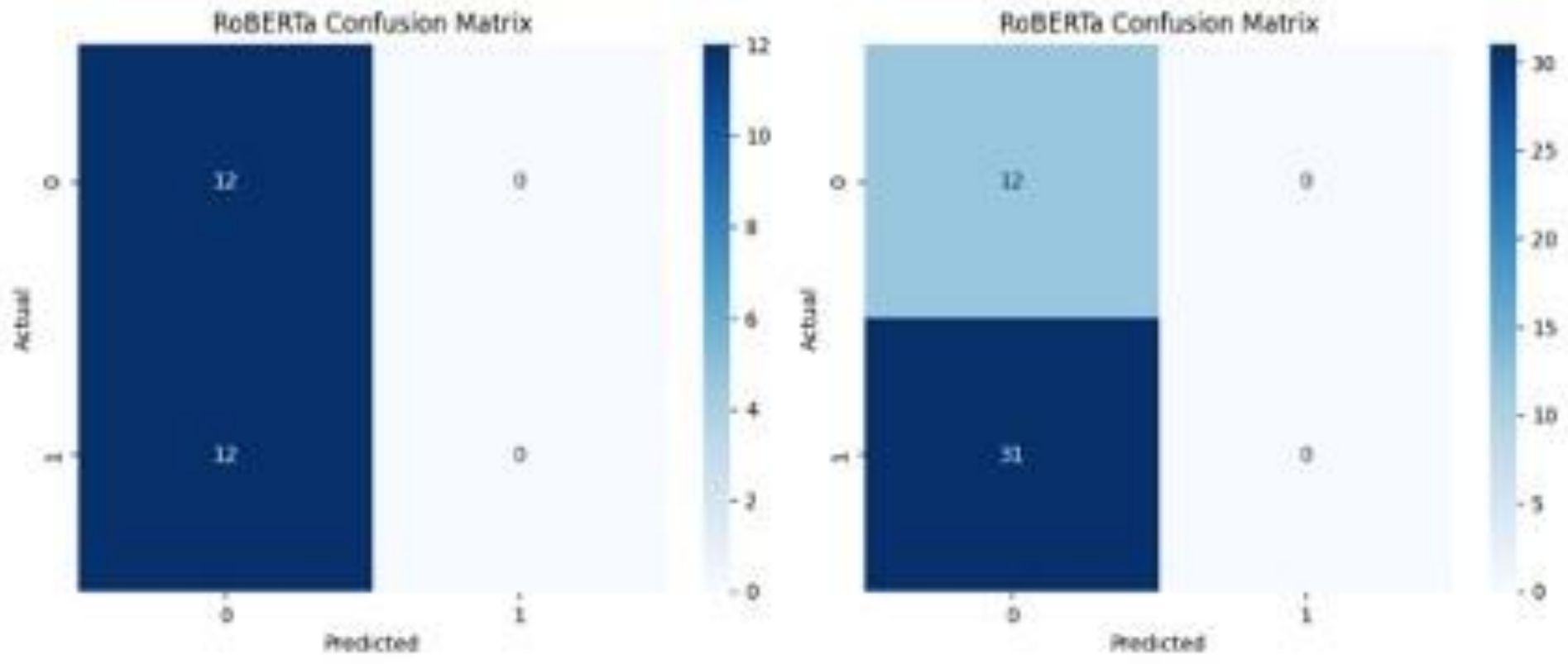

Balanced dataset Imbalanced dataset

*Figure 8: RoBERTa - Confusion Matrix for Binary Sentiment Analysis (Balanced vs. Imbalanced Dataset)*

For the RoBERTa model on the imbalanced dataset, the model shows a significant bias towards predicting the negative class (Class 0), failing to correctly classify any instances as positive (Class 1). The precision and recall for Class 1 are both 0.00, resulting in an overall low F1-score and accuracy. On the balanced dataset, the model performs better in terms of precision and recall for the negative class (Class 0) but fails to correctly classify any instances as positive (Class 1). The precision and recall for Class 1 remain 0.00, with an overall accuracy of 0.50.

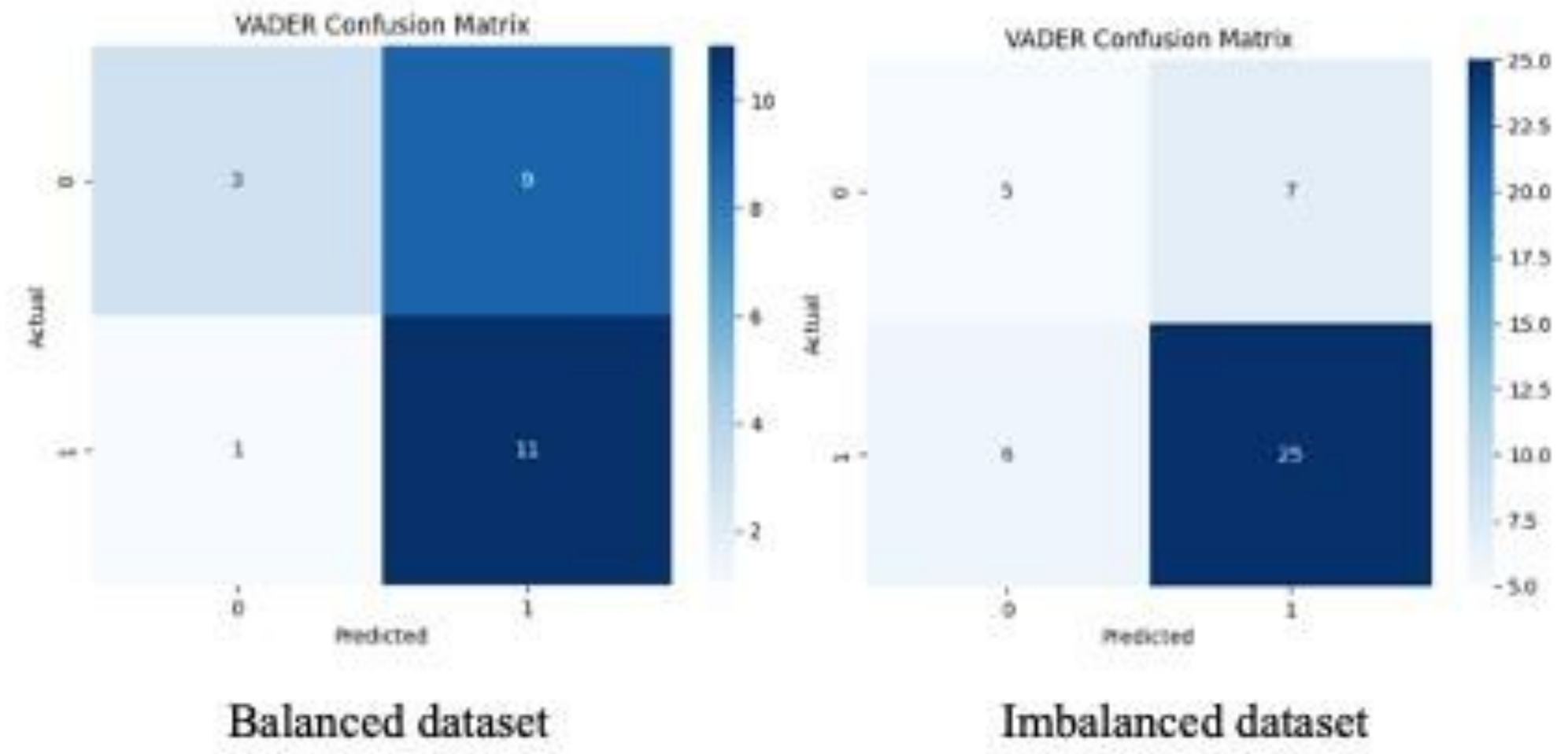

*Figure 9: VADER - Confusion Matrix for Binary Sentiment Analysis (Balanced vs. Imbalanced Dataset)*

For the VADER model on the imbalanced dataset, the model shows a more distributed classification but still struggles with correctly classifying all instances. It performs relatively better in identifying positive instances (Class 1) with higher precision, recall, and F1-score, resulting in an overall accuracy of 0.70. On the balanced dataset, the model shows improved performance, with higher precision and recall for both classes. However, there are still notable misclassifications, resulting in an overall accuracy of 0.58.

4.5.4 Best Model Analysis

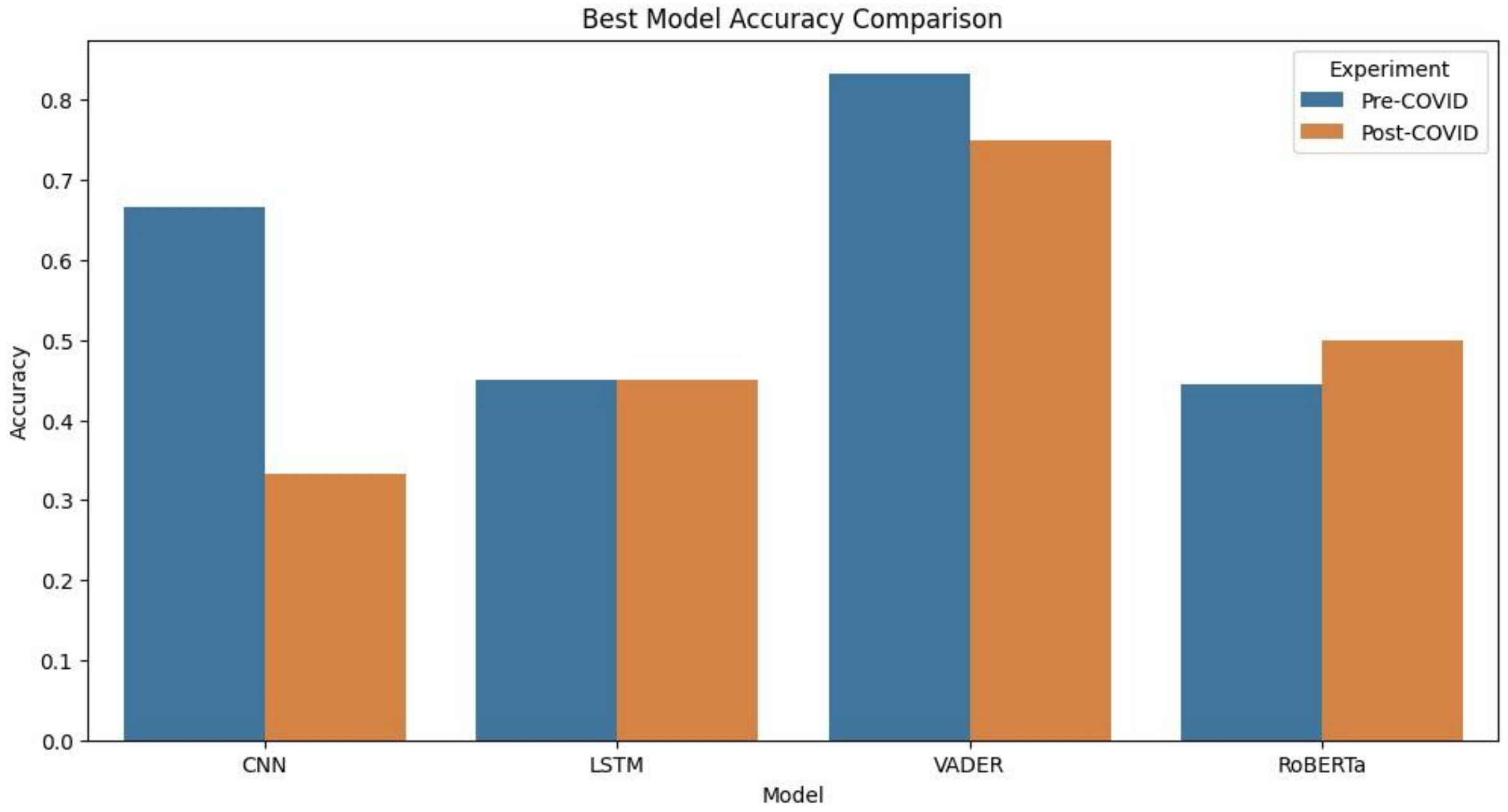

*Figure 10: Model Accuracy Comparison*

In Figure 10, we observe the accuracy comparison among the models (CNN, LSTM, VADER, RoBERTa) pre- and post-COVID. VADER shows the highest accuracy pre-COVID at 0.8333, followed by CNN at 0.6667. Post-COVID, VADER still has the highest accuracy at 0.7500, followed by RoBERTa at 0.5000.

Both CNN and LSTM models experience a significant drop in accuracy postCOVID, with CNN at 0.3333 and LSTM remaining constant at 0.4500. This suggests that VADER is the best performing model in terms of accuracy both pre- and post-COVID, with a notable decrease in performance post-COVID for CNN.

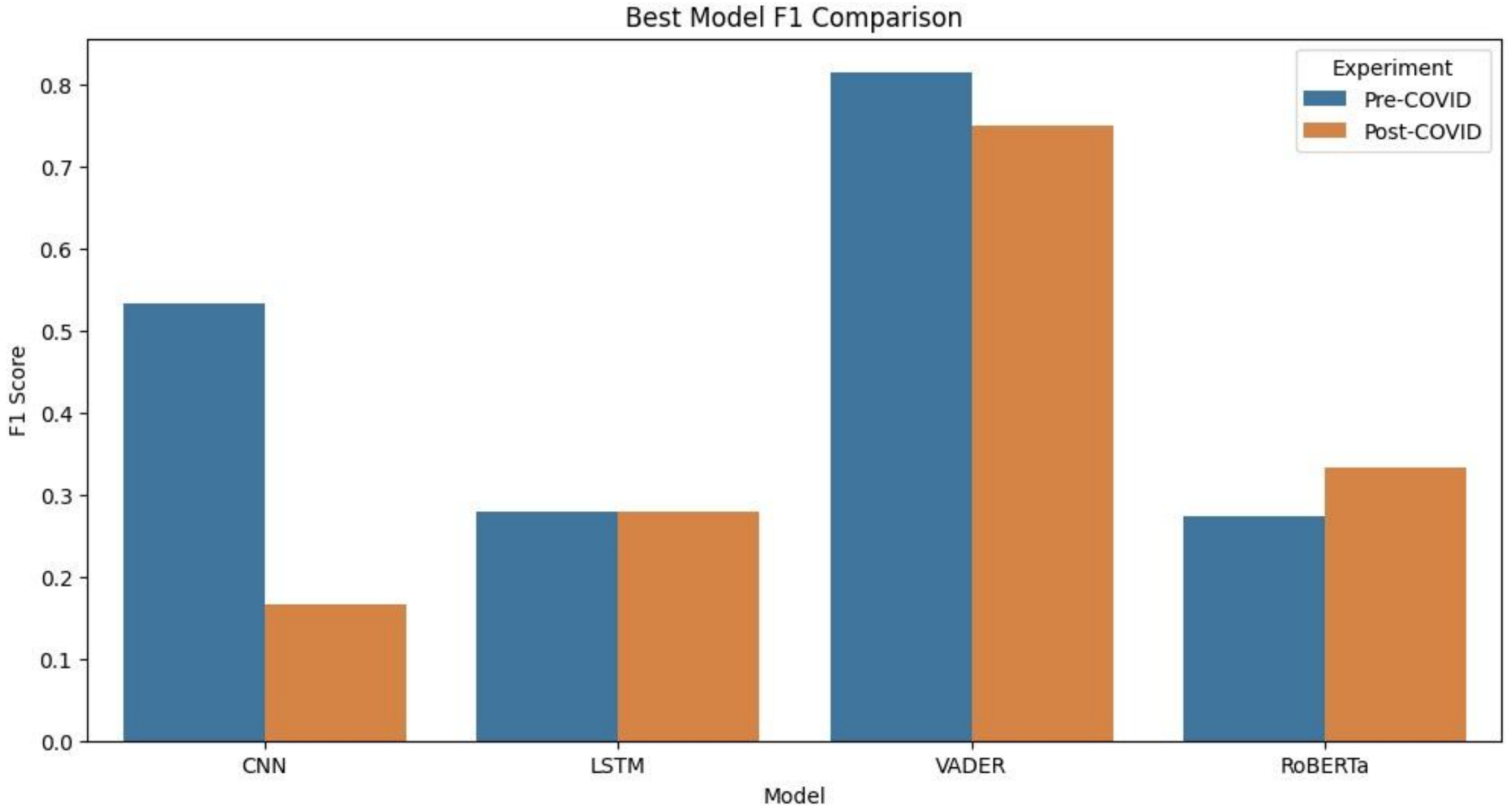

*Figure 11: Best Model F1 Score Comparison*

Figure 11 presents the F1 score comparison among the models. VADER again outperforms other models with the highest F1 score pre-COVID at 0.8148 and postCOVID at 0.7500. CNN shows a significant drop from 0.5333 pre-COVID to 0.1667 post-COVID. LSTM has a constant F1 score of 0.2793 pre- and postCOVID, indicating no change in its performance. RoBERTa shows an improvement in its F1 score from 0.2735 pre-COVID to 0.3333 post-COVID. This suggests that VADER is the best overall performer in terms of F1 score, effectively balancing precision and recall, while CNN's performance significantly deteriorates post-COVID.

4.5.5 VADER – Model application

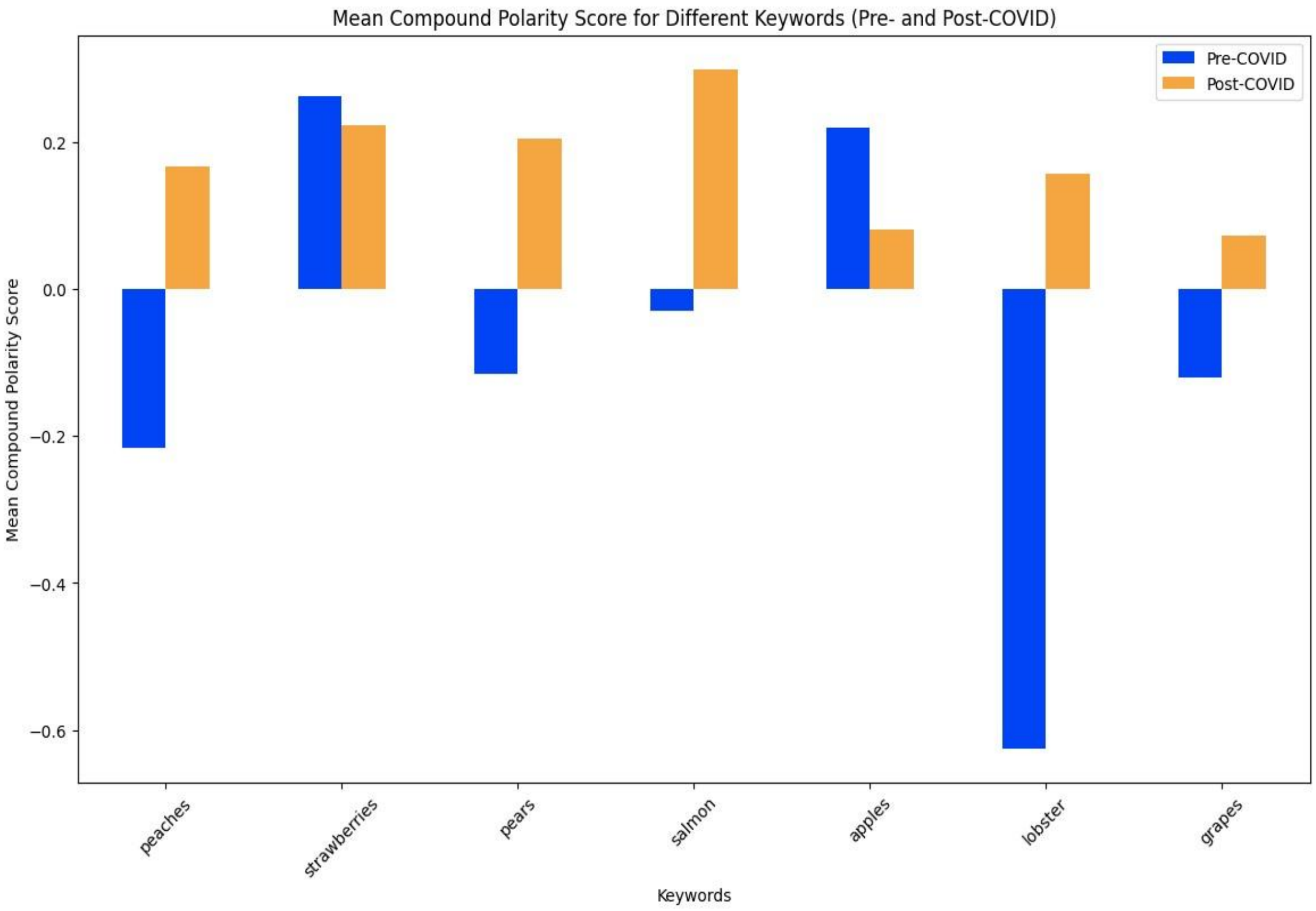

*Figure 12: Mean Compound Polarity Score for Keywords (Pre- and Post-Covid)*

In figure 12, the mean compound polarity scores for various keywords are compared between the pre-COVID and post-COVID periods. The polarity scores for keywords such as peaches, pears, salmon, lobster, and grapes show a noticeable shift from negative to positive after COVID-19. For example, "peaches" increased from -0.215 to 0.168 and "salmon" from -0.030 to 0.300. This indicates a shift in sentiment towards these keywords from negative to positive. Conversely, "strawberries" and "apples" experienced a slight decline in their positive polarity scores post-COVID, though they remained positive overall. "Lobster" showed a significant shift from a strongly negative polarity of -0.625 to a positive polarity of

0.156, indicating a substantial change in sentiment.

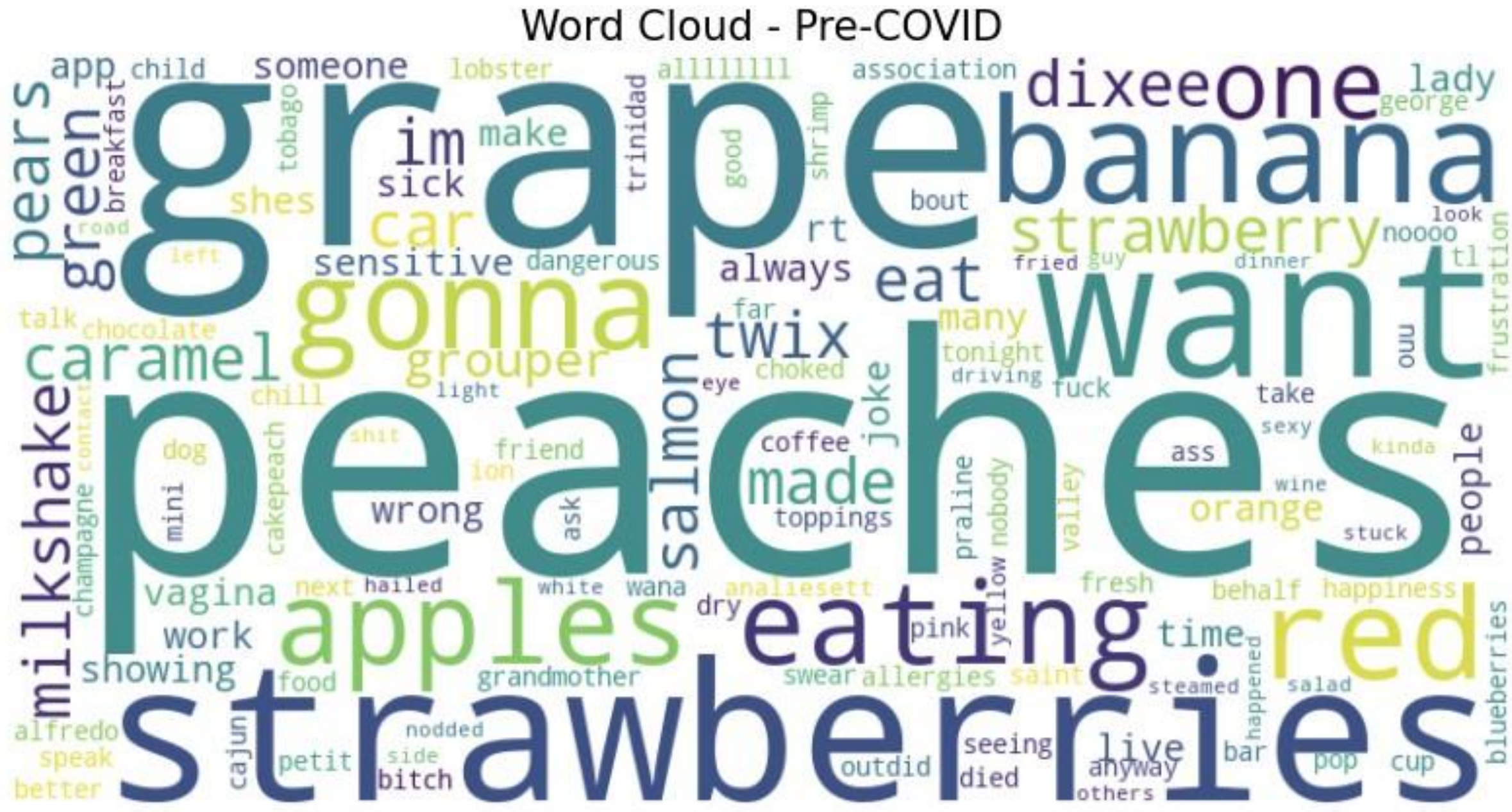

*Figure 13: Word Cloud of Keywords on Pre-COVID Dataset*

The word cloud in figure 13 visualizes the frequency and sentiment of keywords in the pre-COVID dataset. Keywords like "strawberries" and "apples" appear prominently, reflecting their positive sentiment with higher frequency. In contrast, keywords like "lobster" and "peaches" have a more negative sentiment, as shown by their less frequent but still noticeable presence.

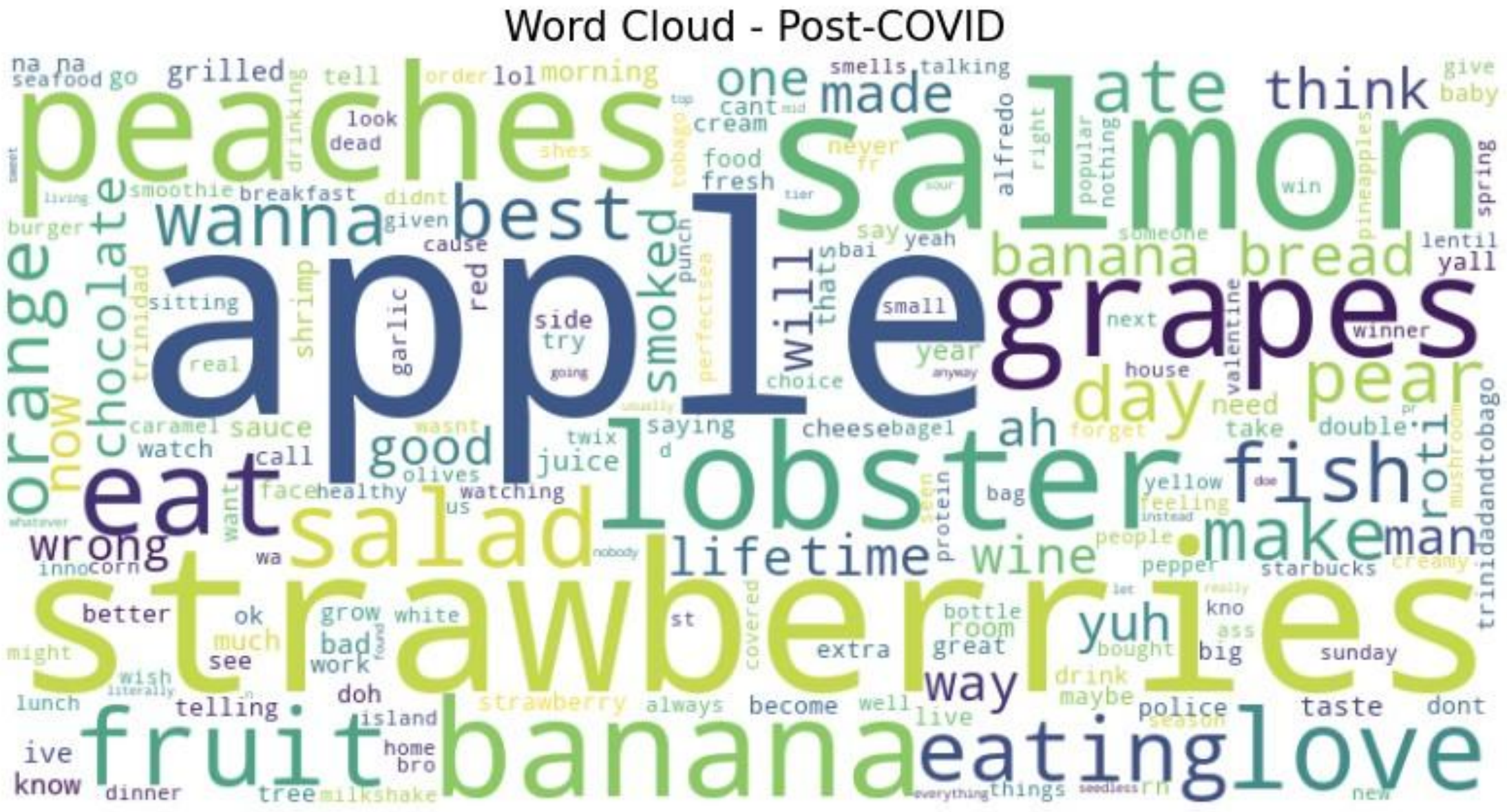

*Figure 14: Word Cloud of Keywords on Post-COVID Dataset*

The word cloud in figure 14 illustrates the frequency and sentiment of keywords in the post-COVID dataset. There is a noticeable increase in the prominence of keywords such as "peaches" and "salmon," which now carry a positive sentiment. The shift in sentiment is visually evident, with previously negative keywords like "lobster" now appearing with a positive sentiment, indicating a change in public perception.

## Section 5: Discussion

### 5.1 Performance Comparison

In response to Research Question 1, the results highlight significant differences in model performance. The CNN model demonstrated relatively stable performance on the imbalanced dataset, achieving an accuracy of 57.14% and an F1 score of 0.4908 with fixed hyperparameters. Notably, varying the hyperparameters did not significantly enhance the model's performance; in fact, it slightly reduced the accuracy and F1 score. When applied to the balanced dataset, the CNN's performance deteriorated, suggesting potential overfitting issues due to the reduced dataset size.

In contrast, the LSTM model exhibited lower overall performance compared to CNN, with an accuracy of 38.10% and a similar F1 score using fixed hyperparameters. Like CNN, hyperparameter variations did not improve the LSTM model's performance. The balanced dataset led to even poorer results, indicating that LSTM struggled more significantly with reduced and balanced data, perhaps due to its inherent complexity and need for more extensive training data. The underperformance of LSTM could be attributed to its reliance on sequential data processing, which may not be as effective with the characteristics of our Twitter dataset.

Among the models, VADER outperformed all others in multi-class sentiment analysis on the imbalanced dataset, achieving an accuracy of 61.90%. VADER also excelled in binary sentiment analysis, with the highest accuracy (69.77%) and F1 score (0.6935), highlighting its robustness and suitability for sentiment analysis tasks. Conversely, RoBERTa, despite its advanced architecture, showed lower performance in both multi-class and binary classifications. It particularly struggled with the balanced dataset, suggesting that while RoBERTa is a powerful model, it may require more sophisticated tuning and larger datasets for optimal performance.
The complexity and need for extensive fine-tuning of RoBERTa might have contributed to its underperformance, indicating a potential area for future research to explore more effective fine-tuning strategies.

### 5.2 Optimal Configurations

Research Question 2 focused on identifying optimal configurations for each model. For CNN, the best hyperparameters were found to include the ReLU activation function, Adam optimizer, a batch size of 16, 10 epochs, and a dropout rate of 0.5. These settings yielded the highest performance metrics, underscoring their effectiveness. Similarly, the LSTM model's optimal configuration involved 64 LSTM units, ReLU activation, Adam optimizer, a batch size of 16, 10 epochs, and a dropout rate of 0.5. VADER's performance was maximized with a sentiment classification threshold of 0.05. For RoBERTa, using the base pre-trained model with 3 fine-tuning epochs and a learning rate of 2e-5 provided the best results, albeit still underperforming compared to VADER and CNN.

VADER's performance benefits stemmed from its inherent design focused on sentiment lexicons, proving effective in scenarios involving imbalanced datasets and achieving competitive accuracy metrics. In contrast, RoBERTa's performance highlighted the importance of rigorous pre-training and fine-tuning processes, with room for improvement in adapting to specific dataset characteristics and optimizing hyperparameters for enhanced sentiment analysis. The experiments underscore the significant impact of data imbalance and the critical role of hyperparameter tuning on model performance. Generally, models performed better on imbalanced datasets, possibly due to the larger size and diversity of the data. Hyperparameter tuning had varied effects; for some models like CNN, it led to minor improvements, whereas for LSTM, it had little to no impact. Balancing the dataset generally reduced performance, potentially due to overfitting on the smaller, balanced datasets.

### 5.3 Impact of COVID-19

The comparative analysis of sentiment trends before and after the COVID-19 pandemic revealed notable shifts. The performance of all models declined postCOVID, suggesting changes in public sentiment and tweet characteristics. VADER maintained the highest accuracy both pre- and post-COVID, indicating its robustness in handling dynamic changes in sentiment expressions over time. The decrease in performance for CNN and LSTM post-COVID suggests these models are more sensitive to shifts in data distribution and sentiment patterns, possibly due to their reliance on specific data characteristics that may have changed during the pandemic.

The mean compound polarity scores for various keywords showed noticeable shifts from negative to positive after COVID-19, indicating a positive re-evaluation of these food items during the post-COVID period. Keywords such as “strawberry” and “apples” showed slight declines in their positive polarity scores post-COVID, indicating minor negative shifts in sentiment. Conversely, keywords like "lobster" and "peaches" had more negative sentiment pre-COVID but saw increased prominence and positive sentiment post-COVID, suggesting a change in public perception of these specific food items over time.

### 5.4 Practical Implications

In addressing Research Question 3, the optimized sentiment analysis models have significant potential for real-time applications in monitoring public sentiment regarding imported food items in Trinidad and Tobago. Integrating these models into live systems can facilitate continuous sentiment tracking, providing stakeholders with timely insights into consumer opinions and concerns. Policymakers can leverage these insights to adjust import strategies proactively, addressing public dissatisfaction with certain imported food products. For example, the VADER model, with its high accuracy and F1 score, could be integrated into systems to monitor and analyse sentiment trends, helping to manage the import bill by identifying and addressing negative sentiments towards specific imported items. By using these models, stakeholders can identify trends and shifts in sentiment that may signal emerging issues or opportunities in the import market. This proactive approach can lead to more informed decision-making, potentially optimizing the import bill and enhancing consumer satisfaction. Furthermore, real-time sentiment analysis can help detect negative trends early, allowing for swift interventions to mitigate any adverse impacts on public perception and import dynamics.

### 5.5 Limitations

One significant limitation encountered in this study was related to the collection of tweets. Due to recent changes in Twitter's policies and the transition to what is now called X, accessing tweets through the Twitter developer platform required a paid subscription. This restriction posed a challenge, as third-party tweet-collecting websites were no longer operational. To overcome this, a custom HTML code was developed to collect tweets, which added complexity and potential limitations to the data collection process. This constraint might have influenced the sample size and diversity of the collected tweets, potentially impacting the robustness of the analysis.

## Section 6: Conclusion

This study explored the effectiveness of CNN, LSTM, VADER, and RoBERTa models for sentiment analysis on Twitter data about imported food items in Trinidad and Tobago. The findings showed that when it came to model performance, VADER emerged as the best-performing model overall, particularly excelling in binary sentiment analysis and demonstrating robustness in varying data conditions and time periods. CNN showed moderate performance but struggled with balanced datasets and post-COVID data shifts. LSTM consistently underperformed across most scenarios, indicating a need for alternative architectures or more extensive tuning. RoBERTa, despite its advanced architecture, did not perform as well, suggesting it requires more data and finetuning for optimal performance. For this study the optimal configurations for CNN and LSTM included using the Adam optimizer, a batch size of 16, and 10 epochs. For VADER, a threshold of 0.05 was effective, while RoBERTa performed best with a learning rate of 2e-5 and 3 fine-tuning epochs. The practical implications of these findings are significant, suggesting that optimized sentiment analysis models can provide valuable real-time insights into public sentiment. This capability can enhance decision-making processes related to import policies and strategies by aiding policy makers and stakeholders in making informed decisions, ultimately contributing to more efficient management of the import bill and improved consumer satisfaction. By leveraging real-time sentiment analysis, stakeholders can anticipate and respond to public sentiment shifts promptly, thereby optimizing import strategies and potentially reducing import bills.

Future research should explore hybrid approaches that combine the strengths of different models and more advanced tuning techniques to further enhance performance, especially for models like LSTM and RoBERTa which showed room for improvement, as well as extending the dataset to include a broader range of sentiments and more diverse topics. Additionally, examining the long-term trends in sentiment and their impacts on import dynamics over multiple years would provide deeper insights and strengthen the external validity of the findings. The integration of sentiment analysis into broader socio-economic studies could further elucidate the relationship between public sentiment and market behaviours, offering more comprehensive strategies for managing import-related challenges. This thesis advocates for the integration of advanced ML techniques into governance frameworks, promoting adaptive and responsive decision-making aligned with public sentiment dynamics. In essence, this study not only advances the understanding of sentiment analysis methodologies but also provides a foundation for future research and practical applications in leveraging AI-driven insights for informed policy-making and governance. With respect to the limitation on data collection, although the workaround while necessary, it may have affected the sample size and diversity of the dataset. Future research should consider these constraints and explore alternative data collection methods or secure adequate funding to access comprehensive data through official channels.

**Appendicies**

**Appendix 1: Custom Web Scrapping Tool**

```
from bs4 import BeautifulSoup import os import json

tweet_path = "/Users/cassandradaniels/Desktop/Project/tweets"

for root, dirs, files in os.walk(tweet_path):     for file in sorted(files):     if file[-4:]=="html":     with open(f"{tweet_path}/{file}") as htmlfile:     tweet_soup = BeautifulSoup(htmlfile,'lxml')     tweet_text = tweet_soup.find("div", {"class":'css-1rynq56 r-bcqeeo rqvutc0 r-37j5jr r-1inkyih r-16dba41 r-bnwqim r-135wba7'}).text
```

```
tweet_time = tweet_soup.find('a', {'class':'css-1qaijid r-bcqeeo r-qvutc0 r-poiln3 r-xoduu5 r-1q142lx r-1w6e6rj r-9aw3ui r-3s2u2q r-1loqt21'}).text                tweet_views = tweet_soup.find('div', {'class':'css-175oi2r r-xoduu5 r-1udh08x'}).text                new_data={
        'text': tweet_text,
        'Date/Time': tweet_time,
        'Views': tweet_views,
        'Location' : tweet_location,
        'Username' : tweet_user,
        'Replies' : reply_count,
        'Likes' : like_count

    }
    with open(f'{tweet_path}/further_user_data.json', 'a') as further_data:
        json.dump(new_data, further_data, indent=4)
```

**Appendix 2: Examples of Classified Tweets**

Positive Sentiment Examples**:**

1. " strawberry and banana smoothie for the win"
2. " bananas are a major key fr fr"

Neutral Sentiment Examples:

1. " i picked a pair of pears with my peers"
2. " wym peaches was right down the road"

Negative Sentiment Examples:

1. " is seventyfive cents for oneeeee bobby na na na na forget the lobster roti we concerned about the wrong things cause wa yuh mean to say "
2. " bought doubles for breakfast this morning when ik i shouldve bought a couple apples or pears inno i could dead rn"